\documentclass[10pt,twocolumn,letterpaper]{article}
\usepackage{CVPR_2025}              % To produce the CAMERA-READY version

\usepackage{multirow}
\usepackage{algorithm}
\usepackage{algorithmic}
\usepackage{amsmath}
\usepackage{amsthm}

\definecolor{cvprblue}{rgb}{0.21,0.49,0.74}
\usepackage[pagebackref,breaklinks,colorlinks,allcolors=cvprblue]{hyperref}

\usepackage{subfigure}

\newcommand{\calA}{\mathcal{A}}

\newcommand{\calU}{\mathcal{U}}

\newcommand{\calI}{\mathcal{I}}

\newcommand{\calN}{\mathcal{N}}

\newcommand{\VV}{\mathbb{V}}
\newcommand{\QQ}{\mathbb{Q}}

\newcommand{\gray}[1]{{\color{gray}#1}}

\newtheorem{theorem}{Theorem}

\newtheorem{rmk}{Remark}
\newtheorem{definition}{Definition}

\setlist[itemize]{leftmargin=*}
\usepackage{soul}

\title{FedMIA: An Effective Membership Inference Attack Exploiting  \\``All for One" Principle in Federated Learning}

\author{
Gongxi Zhu$^1$ \quad
Donghao Li$^2$  \quad
Hanlin Gu$^{2,3}$  \footnotemark[1] \quad 
Yuan Yao$^2$\quad
Lixin Fan$^3$ \quad
Yuxing Han$^1$
\vspace{1mm}\\
$^1$Tsinghua University\quad 
$^2$The Hong Kong University of Science and Technology\quad 
$^3$Webank\quad
\vspace{1mm} \\
{\tt\small gx.zhu@foxmail.com, dlibf@connect.ust.hk, ghltsl123@gmail.com}  \\
}

\begin{document}
\maketitle
\renewcommand{\thefootnote}{\fnsymbol{footnote}} 
\footnotetext[1]{Corresponding author.} 

\begin{abstract}
Federated Learning (FL) is a promising approach for training machine learning models on decentralized data while preserving privacy. However, privacy risks, particularly Membership Inference Attacks (MIAs), which aim to determine whether a specific data point belongs to a target client’s training set, remain a significant concern. Existing methods for implementing MIAs in FL primarily analyze updates from the target client, focusing on metrics such as loss, gradient norm, and gradient difference. However, these methods fail to leverage updates from non-target clients, potentially underutilizing available information.
In this paper, we first formulate a one-tailed likelihood-ratio hypothesis test based on the likelihood of updates from non-target clients. Building upon this formulation, we introduce a three-step Membership Inference Attack (MIA) method, called FedMIA, which follows the "all for one"—leveraging updates from all clients across multiple communication rounds to enhance MIA effectiveness. Both theoretical analysis and extensive experimental results demonstrate that FedMIA outperforms existing MIAs in both classification and generative tasks. Additionally, it can be integrated as an extension to existing methods and is robust against various defense strategies, Non-IID data, and different federated structures. Our code is available in \url{https://github.com/Liar-Mask/FedMIA}.

% Our code is available in \url{https://github.com/Liar-Mask/FedMIA}.

\end{abstract}
\section{Introduction}
\vspace{-5pt}

Federated learning (FL) \citep{mcmahan2016federated, mcmahan2017communication, konevcny2016federated} has emerged as a promising approach for training machine learning models on decentralized data sources while ensuring data privacy. Despite its advantages, the privacy risks associated with the information exchanged during FL have attracted significant research attention. Membership Inference Attacks (MIAs) in FL aim to determine whether a specific data point was part of a particular client’s training dataset, typically performed by adversaries positioned on the server side.
In contrast to Gradient Inversion Attacks (GIAs) \citep{zhu2019deep, geiping2020inverting}, MIAs \citep{nasr2018machine} do not rely on strong assumptions, such as small batch sizes or local training epochs, and thus remain significantly underexplored within the FL context.

\begin{figure}[t]
    \flushleft 
    \includegraphics[scale=0.4]{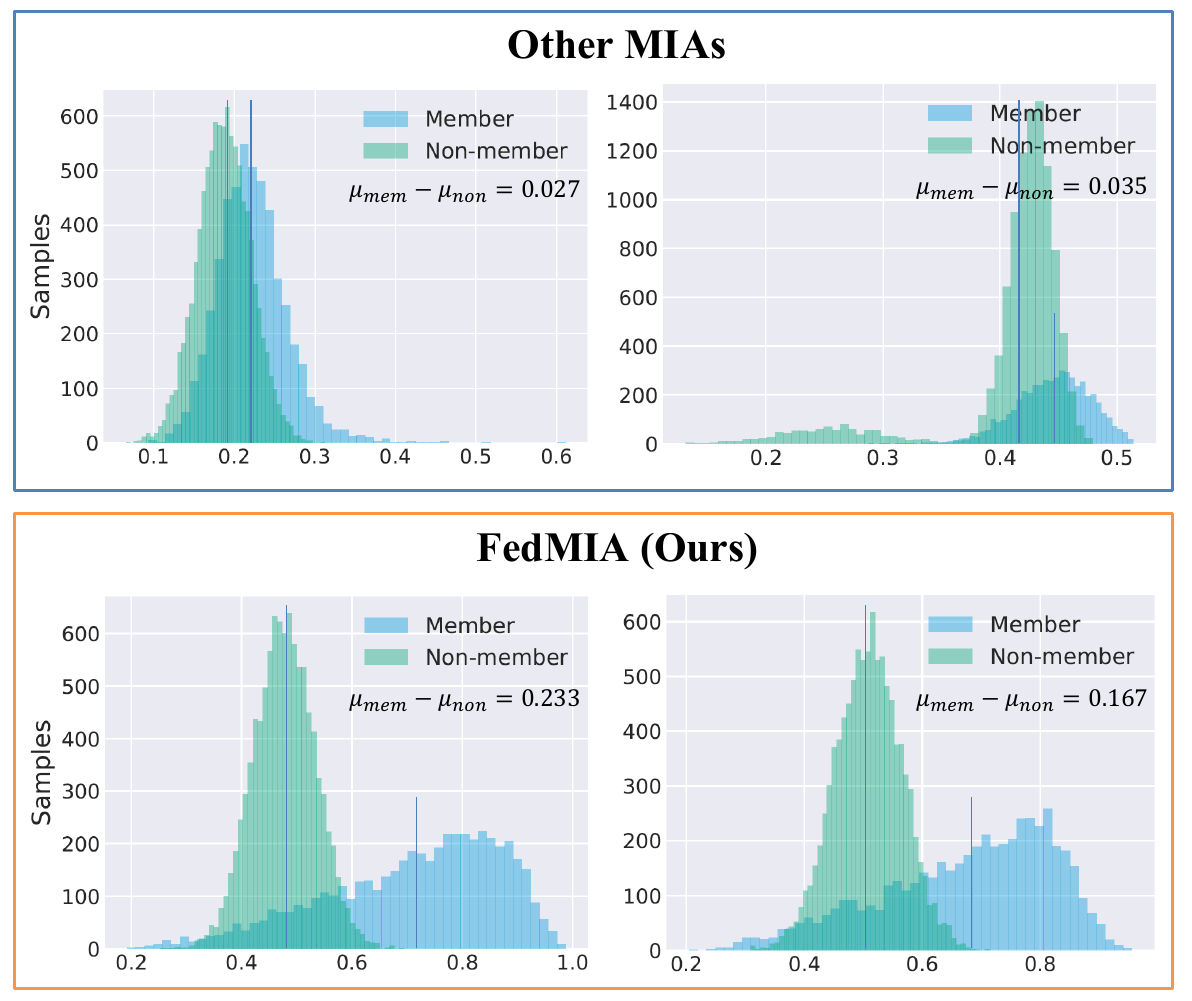}
    \caption{The distributions of member and non-member samples of FedMIA (the second row: FedMIA-I (ours) and FedMIA-II (ours)) and other MIAs (the first row: Grad-Cosine \cite{li2022effective}, Loss-Series \cite{gu2022cs}) on ResNet-CIFAR100. It shows the obvious gap between the mean of the member and non-member ($\mu_{mem}-\mu_{non}$) for the proposed FedMIA compared to other methods. }
    \label{Fig:Intro}
\end{figure}

Most existing MIAs in FL \citep{nasr2019comprehensive, nasr2018machine, zari2021efficient, suri2022subject, zhang2020gan, li2022effective} focus on inferring membership \textit{solely from the updates of the target clients}, utilizing gradient norms, loss values, and gradient differences. However, these methods overlook the valuable information contained in updates uploaded by non-target clients. Recent work \citep{gu2022cs, he2024enhance} attempts to enhance the effectiveness of MIAs by incorporating shadow models into FL. While these methods make use of additional information, they require an auxiliary dataset to train the shadow model, which may not be feasible in the FL setting, as the server does not have access to the private data of local clients. Furthermore, training a shadow model incurs additional computational costs for the adversary.

To address the limitations of existing approaches, this paper proposes an alternative method that leverages updates from non-target clients, denoted as \( I_{\text{non-tar}} \), instead of relying on an auxiliary dataset for Membership Inference Attacks (MIAs). A key challenge in this approach is that the server lacks knowledge of which updates correspond to data trained on the target client's dataset, making it difficult to estimate the distribution of updates trained on the target data, denoted as \( \mathcal{Q}_{\text{in}} \). To overcome this challenge, we demonstrate that it is possible to estimate the distribution of updates that are not trained on the target data, denoted as \( \mathcal{Q}_{\text{out}} \), using updates uploaded by non-target clients \( I_{\text{non-tar}} \). Since each client’s data is disjoint, at least some of the updates from non-target clients \( I_{\text{non-tar}} \) will not be trained on the target data, i.e., \( \exists I_{\text{non-tar}} \sim \mathcal{Q}_{\text{out}} \). Based on this, we formulate a one-tailed likelihood-ratio hypothesis test using the distribution of updates that were not trained on the target data, \( \mathcal{Q}_{\text{out}} \), to perform the MIA.

Building on this one-tailed likelihood-ratio hypothesis test, we introduce a three-step Membership Inference Attack (MIA) method called FedMIA, which follows the ``all for one”—leveraging
updates from all clients across multiple communication rounds. The first step involves computing a low-dimensional representation to simplify the distribution of  updates. In the second step, we estimate the distribution of updates not trained on the target data for each communication round. Finally, we apply the one-tailed likelihood-ratio test based on the estimated distribution of updates not trained on the target data. This test is further extended by incorporating updates from all communication rounds. The proposed FedMIA has three advantages: 1) FedMIA achieves superior performance compared to other MIAs by utilizing the updates information from non-target clients in Sect. \ref{sec:exp}; 2) FedMIA can be integrated into existing methods as an extension in Sect. \ref{subsec: MIA-setup}; and 3) FedMIA is robust against six defense methods, two federated structures, varying degrees of Non-IID data, and different client counts, communication rounds, and local epochs. Our contributions are summarized as the following:
\begin{itemize}
\item We first formulate the non-target updates as a one-tailed likelihood-ratio hypothesis test to evaluate the performance of the updates without being trained on target data. Theorem \ref{thm:1} proves the validity of our formulation.

\item Building on this hypothesis test, we introduce a three-step Membership Inference Attack (MIA) method, called FedMIA, which leverages updates from all clients and communication rounds to enhance MIA effectiveness.

\item Extensive results
show that the proposed FedMIA: 1) achieves superior performance compared to other MIAs in both classification and generative tasks (see Fig. \ref{Fig:Intro}); 2) can be integrated into existing methods as a plug-in; and 3) is robust against six defense methods, two federated structures, varying degrees of Non-IID data, and different client counts, communication rounds, and local epochs.
\end{itemize}

\section{Related work}
\subsection{Federated Learning}
{Federated learning was originally proposed as a collaborative approach for training machine learning models without the need to share private data among multiple parties \citep{mcmahan2016federated,mcmahan2017communication,konevcny2016federated,yang2019federated}. However, more recently, the concept of ``trustworthy federated learning'' has been introduced by \citep{kang2023optimizing}. This variant of federated learning places a heightened emphasis on the preservation of privacy throughout the federated learning process. This shift in focus reflects the increasing awareness of privacy concerns and the recognition of the importance of robust security measures in federated learning systems}

\subsection{Membership Inference Attack }
% \LDH{Too Much?} \HL{Yes, move table here}

MIA is a widely studied privacy attack in centralized learning scenarios. Depending on the information available to the attacker, MIA can be categorized into black-box attack (where only the output of the model can be obtained) \citep{shokri2017membership,ndss19salem,yeom2018privacy,sablayrolles2019white,song2020systematic,choo2020label,hui2021practical,truex2019demystifying} and white-box attack (where the entire model is available) \citep{nasr2019comprehensive,rezaei2020towards}.

In the context of federated learning, Nasr et al. \cite{nasr2019comprehensive} first analyzed membership inference attacks in federated learning and proposed both passive and active attacks. In a passive attack, the attacker solely focuses on obtaining membership leaks based on accessible information without disrupting or compromising the normal training process. Conversely, an active attack involves the ability to modify the updates of federated learning, thereby increasing the vulnerability of the trained models to attacks.
Zari et al. \cite{zari2021efficient} proposed a membership inference attack for federated learning that utilizes the probabilities of correct labels under local models at different epochs for inference. However, this approach requires member samples for auxiliary attacks. Li et al. \cite{li2022effective} proposed a passive membership inference attack that does not require training on member samples. They designed two metric features based on the orthogonality of gradients to distinguish whether a sample is a member. Hu et al.\cite{hu2023source} designed an inference attack  to facilitate an honest-but-curious server to identify the training record's source client, which bases on but extends MIAs to source inference.
Moreover, inspired by work on worst-case privacy auditing, Aerni et al. \cite{aerni2024evaluations} introduced an efficient assessment method that accurately reflects the privacy of defenders at their most vulnerable data points.
% We have provided a table below to compare the differences between our proposed attack and existing attacks:

% \begin{table}[h]
% \centering
% \tiny
% \caption{Comparing different methods  }
% \begin{tabular}{@{}lccccc@{}}
% \toprule
%             % & Setting        & Requirements   & \multicolumn{3}{c}{Information Used}                 \\ \midrule
% Attack name & FL     & Extra data     & Shadow model   & Multiple updates & Multiple clients \\
% \midrule
% Lira\cite{carlini2022membership}        & \XSolidBrush   & Multi & Multi & \XSolidBrush     & \XSolidBrush     \\
% Black-Box \cite{yeom2018privacy}    & Multi & \XSolidBrush   & \XSolidBrush   & \XSolidBrush     & \XSolidBrush     \\
% Grad-Norm \cite{nasr2019comprehensive}    & Multi & \XSolidBrush   & \XSolidBrush   & \XSolidBrush     & \XSolidBrush     \\
% Fed-Loss \cite{li2022effective}    & Multi & \XSolidBrush   & \XSolidBrush   & Multi   & \XSolidBrush     \\
% Cos-Sim  \cite{li2022effective}       & Multi & Multi & \XSolidBrush   & Multi   & \XSolidBrush     \\
% Grad-Diff \cite{li2022effective}   & Multi & Multi & \XSolidBrush   & Multi   & \XSolidBrush     \\
% Ours        & Multi & \XSolidBrush   & Multi & Multi   & Multi   \\ \bottomrule
% \end{tabular}
% \end{table}

% \section{Attack Method: Utilizing the information from Multiple Clients and Training Epochs}
\section{An Effective MIA in FL} \label{sec:MIAs-FL}
In this section, we first present the setting of federated learning (FL) in Sect. \ref{subsec:setting}. Subsequently, we formulate the Membership Inference Attacks (MIAs) in FL as a one-tailed likelihood ratio test in Sect. \ref{subsec: LRT}. Building upon this formulation, we introduce an effective MIA in Sect. \ref{sec:proposed}.

\subsection{Setting} \label{subsec:setting}
\noindent\textbf{Horizontal Federated Learning.}
We consider a \textit{horizontal federated learning} (HFL) \citep{yang2019federated,mcmahan2017communication} setting consisting of one server and $K$ clients. We assume $K$ clients have their local dataset $D_k = \{(x_{k,i},y_{k,i}) \}_{i=1}^{n_k}, k=1\cdots K$, where $x_{k,i}$ is the input data, $y_{k,i}$ is the label, and $n_k$ is the total number of data points for $k_{th}$ client. Since we focus on evaluating membership on each client, \textit{we further assume  $D_k$ are disjoint}. We consider two commonly-used FL frameworks: 1) FedAvg \cite{mcmahan2016federated} that the server aggregates the models uploaded by all clients; 2) FedEmbedding \cite{liang2025diffusion} that the server aggregates the embeddings uploaded by all clients;

% $K$ clients collaboratively train a HFL model $w$ ($F_w$) to optimize the following objective:
% \begin{equation} \label{eq:objective}
%     \min_{w} \sum_{k=1}^K\sum_{i=1}^{n_k}\frac{\ell(F_w(x_{k,i}), y_{k,i})}{n_1+\cdots+n_K},
% \end{equation}
% where $\ell$ is the loss, e.g., the cross-entropy loss. We consider two commonly-used FL frameworks including FedAvg, which uploading the models \cite{mcmahan2016federated}
% We mainly consider two types of 
% This paper addresses both classification and generative tasks, differentiating them by the type of information $I_k^t$ uploaded from various clients. For the classification task, represents the models uploaded by clients \cite{mcmahan2017communication}. In contrast, for the generative task, transferring entire generative models between clients and the server is challenging due to the models' large parameter sizes. Consequently, we consider transferring embeddings (prompts) within the federated learning framework \cite{liang2025diffusion}, as these are significantly smaller than the models.

\noindent\textbf{Threat Model.}
We assume the server are \textit{semi-honest} and do not collude with each other. The server faithfully executes the training protocol but aims to infer the membership information from the specific local clients. 

Specifically, the server implement the attack $\calA$ determine whether a specific sample $(x,y)$ belongs to the target client's dataset $D_{tar}$ based on a series of updates (models, gradients or embeddings) among $K$ clients and $T$ communication rounds: $\calI = \{I_k^t|t\in[T], k\in[K]\}$. However, the server will not actively manipulate these information from the local clients and thus without affecting the utilities of the FL model. The $\calA$ can be represented as the following:
\begin{equation} \label{eq:attack3}
\calA(x,y, \calI) = \begin{cases} 1, & \text{if } (x,y) \in D_{tar} \  \\ 0.& \text{otherwise} \end{cases} 
\end{equation}

\subsection{One-tailed Likelihood-Ratio Test in FL}  \label{subsec: LRT}
In FL, a membership inference attack is to determine the sample $(x,y)$ belonging to target dataset $D_{tar}$ as follows:
\begin{align}
    &H_0: (x,y)\notin D_{tar},    &H_1: (x,y)\in D_{tar}. 
\end{align}
Moreover, for each communication round, the server observes $K$ updates $\{I_k^t\}_{k=1}^K$. According to the updates of the target client $I_{tar}^t$, it is thus natural to see a membership inference attack as performing to guess whether updates $I_{tar}^t$ is trained on the data $(x,y)$ or not. The Likelihood-ratio Test \cite{carlini2022membership} can be represented as:
\begin{equation}
    \Lambda(I_{tar};x,y) = \frac{p_{in}(I=I_{tar}|x,y)}{p_{out}(I=I_{tar}|x,y)},
\end{equation}
where $\QQ_{in}(I|x, y)$ and $\QQ_{out}(I|x, y)$ denote the distribution of updates trained on datasets with and without $(x, y)$, respectively, and $p_{in}$ and $p_{out}$ represents the probability density function of $\QQ_{in}(I|x, y)$ and $\QQ_{out}(I|x, y)$. 

However, \textit{estimating $ \QQ_{in}(x, y)$ in federated learning (FL) presents a challenge, as the attacker (e.g., the server) lacks knowledge of which updates are trained on $(x, y)$.} Therefore, we build the one-tailed Likelihood-Ratio Test using distribution $\QQ_{out}(I|x, y)$ as:
\begin{equation}\label{eq:lam1}
\begin{split}
       \hat\Lambda(I_{tar},x,y) & =
       \sum_{I'<I_{tar}}p_{out}(I=I'|x,y),    
\end{split}
\end{equation}
which is the probability of
observing a confidence as high as the target updates under the null-hypothesis that the target point $(x, y)$ is a non-member. 
\begin{rmk}
It is noted that, given the data sample $(x, y)$, and considering that clients' training datasets are disjoint, at least $K - 2$ updates (except the target updates) are guaranteed not to be trained on $(x, y)$. This enables the estimation of $\QQ_{out}(x, y)$.
\end{rmk}
\begin{rmk} \label{rmk2}
Eq. \eqref{eq:lam1} assumes that the member corresponds to a large value of \( I \). If the member corresponds to a smaller value of \( I \), Eq. \eqref{eq:lam1} is the probability of
observing a confidence as high as the target updates. 
\end{rmk}

Furthermore, when the server observes the $\calI = \{I_k^t|t\in[T], k\in[K]\}$ during $T$ communication rounds, we can extend Eq. \eqref{eq:lam1} as the following by utilizing the temporal information:
\begin{equation}\label{eq:lam2}
\begin{split}
    \Tilde\Lambda(\{I_{tar}^t\}_{t=1}^T,x,y) &= \frac{1}{T}\sum_{t=1}^T\hat\Lambda(I_{tar}^t,x,y) \\
    &= \frac{1}{T}\sum_{t=1}^T\sum_{I'<I_{tar}^t}p_{out}(I=I'|x,y),
\end{split}
\end{equation}
where $\QQ^t_{out}(I|x, y)$ are the distribution of updates trained on datasets without $(x, y)$ in the $t_{th}$ communication round. We establish the validity of Eq. \eqref{eq:lam2} in Theorem \ref{thm:1}, which demonstrates that for all $T$ communication rounds, any member inferred by $\Tilde\Lambda$ is also inferred at least once by $\hat{\Lambda}^t, t\in[T]$.  Furthermore, the worst-case membership leakage occurs when the target sample is inferred as a member in at least one communication round (see proof in Appendix C).
\begin{theorem}\label{thm:1}
Given the threshold $\delta$, let $\VV_t$ be the member sets estimated by $\hat\Lambda^t$  and $\delta$ in communication round $t$. Let $\Tilde\VV$ be the member sets estimated by $\Tilde\Lambda$ and $\delta$. Then we have
\begin{equation}
    \Tilde\VV  \subset (\VV_1 \cup \cdots \cup \VV_T).
\end{equation}
\end{theorem}    
% \begin{rmk}
% We consider the easiest attack samples in the Sect. \ref{}.
% \end{rmk}
% Moreover, when $FPR = 0.1\%$, we have
% \begin{equation}
%     \Tilde{TPR} \leq \sum_{t=1}^T TPR_t.
% \end{equation}

\subsection{The Proposed Method: FedMIA}  \label{sec:proposed}

\begin{figure*}
\vspace{-10pt}
    \centering
    \includegraphics[scale=0.55]{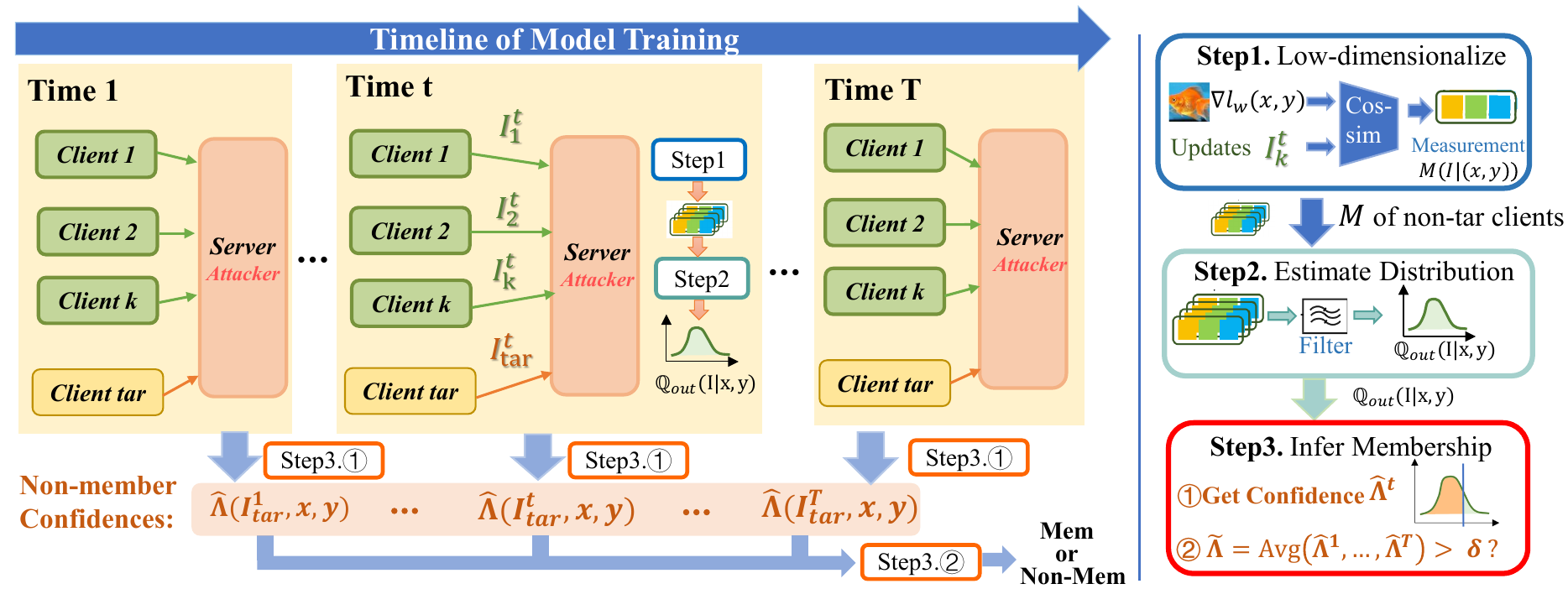}
    \vspace{-10pt}
    \caption{Overview of FedMIA including three steps: 1) Computing the low-dimensional measurement; 2) Estimating the distribution of updates without being trained on target data; 3) Building the one-tailed LRT test and Inferring the membership.}
    \label{Fig:framwork}
% \vspace{-10pt}
\end{figure*}

{Based on the one-tailed likelihood-ration test illustrated in Sect. \ref{subsec: LRT}, we propose a three-step Membership inference attacks by leveraging the spatial and temporal information.}

% \HL{Previous works have not fully utilized the information from all clients and communication rounds \cite{yeom2018privacy,nasr2019comprehensive,li2022effective}}.
% Our goal is to build an MIA framework that utilizes information from multiple communication rounds and multiple clients. To accomplish this, we introduce a three-step MIA framework. 
\vspace{4pt}
\noindent\colorbox{rgb:red!2,65;green!30,60;blue!20,125}{
		\parbox{0.45\textwidth}{\textbf{Step 1: Computing the low-dimensional measurement $M(I|(x,y))$.}}}

Since the updates are high-dimensional, directly estimating $\QQ_{out}(I|x,y)$ is challenging. To address this, we utilize gradient similarity \cite{li2022effective} to map the updates $I(x,y)$ to a low-dimensional variable $M(I|(x,y))$, which allows for an estimation of the distribution as follows:

\begin{equation} \label{eq:MDM}
    M(I|(x,y)) =\displaystyle\frac{ \langle{I ,\frac{ {\partial \ell(\omega,x,y)}}{{\partial \omega}}\rangle }}{{\lVert I \rVert  \lVert \frac{{\partial \ell(\omega,x,y)}}{{\partial \omega}} \rVert}},
\end{equation}
Eq. \eqref{eq:MDM} evaluates the similarity between the uploaded gradient $I$ ($\nabla F$) and the model gradient on the target data $\frac{ {\partial \ell(\omega,x,y)}}{{\partial \omega}}$, with similarity increasing when the target data is a member. The model $\omega$ is the global model.
\begin{rmk}
Eq. \eqref{eq:MDM} represents one approach; other measurements, such as loss or gradient norm \cite{nasr2018machine,gu2022cs}, can also be used. This suggests that our method can be integrated with existing methods that employ different measurements. In Sect. \ref{sec:exp}, we also consider the model loss $\ell(\omega,x,y)$ on target data \cite{gu2022cs} as one measurement.
\end{rmk}
\vspace{4pt}
\noindent\colorbox{rgb:red!2,65;green!30,90;blue!20,125}{
		\parbox{0.45\textwidth}{\textbf{Step 2: Estimating the $\QQ_{out}(M(I|x, y))$ for each communication round $t$.}}}

We first leverage the non-target clients' updates $\{I_k^t| k\in [K] , k \neq tar\}$ to estimate the distribution of $M(I|(x,y)$ without trained on $(x,y)$, i.e., $\Tilde\QQ_{out}(M(I|(x,y)))$. We assume $\QQ_{out}(M(I|x, y))$ is a Gaussian distribution, i.e., $\QQ_{out}(M(I|x, y)) \sim \calN(\mu_{out}, \sigma^2_{out})$. 

If $(x,y)$ is trained on $I_k^t$, then $M(I_k^t|x, y)$ becomes large. Therefore, if $M(I_k | (x, y))$ is exceptionally high for all non-target clients' measurements, the updates from the non-target client $k$ are likely trained on $(x, y)$ with high probability. Consequently, we remove the extreme large values of $M(I_k^t|x, y)$, where $k \neq \text{tar}$, to better estimate $\Tilde\QQ_{out}(M(I|(x,y)))$. To filter the updates sets, we apply the 3-$\sigma$ rule to the updates trained on $(x,y)$. Specifically, we remove the $k$-th update if the following condition holds:

\begin{equation} \label{eq:select}
    M(I_k^t|x, y) > \mu^t + 3\sigma^t,
\end{equation}
\begin{equation}\text{where }
    \begin{cases}
        \mu^t = \frac{1}{K-1}\sum_{j \neq \text{tar}} M(I_j^t|x, y) \\ \sigma^t = \sqrt{ \frac{1}{K-1} \sum_{j \neq \text{tar}} (M(I_j^t|x, y) - \mu^t)^2}.
    \end{cases}
\end{equation}

After filtering, we obtain the update set $\mathcal{U}_t$ which excludes updates trained on the target data $(x, y)$ for communication round $t$ with high probability. Therefore, we estimate the distribution mean and variance of $\mathcal{N}(\mu_{out}^t, v_{out}^t)$ as:
\begin{equation}
\begin{cases}
    \mu^t_{out} = \frac{1}{|\mathcal{U}^t|}\sum_{j \in \mathcal{U}^t} M(I_j^t|x, y), \text{ and}\\
    v_{out}^t = \frac{1}{|\mathcal{U}^t|}\sum_{j \in \mathcal{U}^t} (M(I_j^t|x, y) - \mu^t_{out})^2.
    \end{cases}
\end{equation}

\vspace{1pt}
\noindent\colorbox{rgb:red!30,155;green!20,20;blue!20,30}{\parbox{0.45\textwidth}{\textbf{Step 3: Inferring the membership based on $\Tilde\Lambda(\{I_{tar}^t\}_{t=1}^T,x,y)$.}}}
According the $\mu^t_{out}$, and $ v^t_{out}$ estimated in step 2, we can calculate the $\hat\Lambda^t(I_{tar},x,y)$ of Eq. \eqref{eq:lam1} as:
\begin{equation} \label{eq:lam1-method}
\hat\Lambda(I_{tar}^t,x,y) =\int_{-\infty}^{M(I_{tar}^t|x, y)} \frac{1}{\sqrt{2\pi v^t_{out}}} e^{-\frac{(x- \mu^t_{out})^2}{2v^t_{out}}} dx,
\end{equation}
which is the probability of observing a confidence as low as the target updates under the null-hypothesis that the target point $(x, y)$ is a non-member. Specifically, the small  $\hat\Lambda(I_{tar}^t,x,y)$ indicates the target data is non-member.

Moreover, we utilize the updates of all communication rounds to obtain the $\Tilde\Lambda(I_{tar},x,y)$ of Eq. \eqref{eq:lam2} as:
\begin{equation}\label{eq:lam2-method}
\begin{split}
       &\Tilde\Lambda(\{I_{tar}^t\}_{t=1}^T,x,y) = \frac{1}{T}\sum_{t=1}^T\hat\Lambda(I_{tar}^t,x,y)\\
       & = \frac{1}{T}\sum_{t=1}^T\int_{-\infty}^{M(I_{tar}^t|x, y)} \frac{1}{\sqrt{2\pi v^t_{out}}} e^{-\frac{(x- \mu^t_{out})^2}{2v^t_{out}}},
\end{split}
\end{equation}
Finally, given the threshold $\delta$, the member is determined if $\Tilde\Lambda > \delta$, otherwise is non-member. 

% \noindent\textbf{Step 4: Aggregating p-values from multiple communication rounds.} 
% In the previous step, we demonstrated the estimation of Type-I error rate from a single round. 
% %In federated learning, incorporating multiple rounds of synchronization and aggregating information can further enhance the attack effectiveness. 
% Based on multiple testing theory, we estimate the probability of at least one test being wrong, which can be seen as family-wise error rate \citep{toothaker1993multiple}: 
% \begin{align}\label{eq:aggre}
%     Pr\left(\bigcup_{t=1}^{T} A_j \right) \leq \sum_{t=1}^{T} Pr(M(x,y,w_k^t)>\mu_k ),
% \end{align}
% where event $A_j$ represents the incorrect rejection of the null hypothesis in the $j$-th round, $\bigcup$ denotes the union operation and we use Boole's inequality to give an estimation of its upper bound.

\begin{algorithm}[htbp]
\begin{algorithmic}[1]
\caption{FedMIA}
\STATE \textbf{Input:} Target data sample $(x, y)$ of client $k$, communication rounds $T$, set of client models $\{I_k^t|t \in [T], k \in [K]\}$, a threshold $\delta$.
\STATE \textbf{Output:} membership prediction (0 or 1)
% \STATE $p\_values \leftarrow []$
\colorbox{rgb:red!2,65;green!30,60;blue!20,125}{
		\parbox{0.42\textwidth}{\vbox{\STATE \gray{$\triangleright$ \textit{Computing the low-dimensional measurement}}
\STATE Calculate $M( I_k^t|x, y)$ according to Eq. \eqref{eq:MDM} for all $k\in[K], t\in[T]$;}}}
\colorbox{rgb:red!2,65;green!30,90;blue!20,125}{
		\parbox{0.42\textwidth}{\vbox{\STATE \gray{$\triangleright$ \textit{Estimating the $\QQ_{out}(I|x, y)$}}
\STATE Choose $\calU_t$ according to Eq. \eqref{eq:select};
\STATE     $\mu^t_{out} = \frac{1}{|\mathcal{U}^t|}\sum_{j \in \mathcal{U}^t} M(I_j^t|x, y)$;
\STATE 
$    v_{out}^t = \frac{1}{|\mathcal{U}^t|}\sum_{j \in \mathcal{U}^t} (M(I_j^t|x, y) - \mu^t_{out})^2$;}}}
\colorbox{rgb:red!30,155;green!20,20;blue!20,30}{\parbox{0.42\textwidth}{\vbox{\STATE \gray{$\triangleright$ \textit{Inferring the membership}}
\STATE Calculate $\hat\Lambda(I_{tar}^t,x,y) $ according to Eq. \eqref{eq:lam1-method};
\STATE Calculate $\Tilde\Lambda(\{I_{tar}^t\}_{t=1}^T,x,y)$ according to Eq. \eqref{eq:lam2-method};
\IF{$\Tilde\Lambda(\{I_{tar}^t\}_{t=1}^T,x,y) > \delta$}
    \RETURN 1
\ELSE
    \RETURN 0
\ENDIF}}}
\end{algorithmic}
\end{algorithm}

\begin{table*}[ht] 

\centering
\caption{Comparison of our attack with various MIAs methods on classification tasks and generative tasks. The larger TPR(\%)@FPR=0.1\% and AUC indicates the better attack effectiveness.}
\vspace{4pt}
\small
\label{tab:MIA}
\setlength{\tabcolsep}{3.3pt} % Reducing the column space to avoid large gaps
\begin{tabular}{@{}cc|cccccc|cc@{}}
\toprule
\multicolumn{2}{c|}{MIA Methods}                                                                                                 & \begin{tabular}[c]{@{}c@{}}Blackbox-Loss \\ \cite{yeom2018privacy}\end{tabular} & \begin{tabular}[c]{@{}c@{}}Grad-Cosine \\ \cite{li2022effective}\end{tabular} & \begin{tabular}[c]{@{}c@{}}Grad-Norm \\ \cite{nasr2018machine}\end{tabular} & \begin{tabular}[c]{@{}c@{}}Loss-Series \\ \cite{gu2022cs}\end{tabular} & \begin{tabular}[c]{@{}c@{}}Avg-Cosine \\ \cite{li2022effective}\end{tabular} & \begin{tabular}[c]{@{}c@{}}Grad-Diff \\ \cite{li2022effective}\end{tabular} & \begin{tabular}[c]{@{}c@{}}FedMIA-I \\ (Ours) \end{tabular}& \begin{tabular}[c]{@{}c@{}}FedMIA-II \\ (Ours) \end{tabular} \\ \midrule
\multicolumn{1}{c|}{\multirow{2}{*}{\begin{tabular}[c]{@{}c@{}}AlexNet \\ CIFAR100\end{tabular}}} & TPR & 0.18$\pm$0.05     & 7.26$\pm$0.25   & 0.14$\pm$0.03 & 25.3$\pm$0.88  & 54.66$\pm$1.22 & 20.52$\pm$0.45 & 53.78$\pm$1.24 & \textbf{66.98$\pm$1.74} \\
\multicolumn{1}{c|}{} & AUC & 0.58$\pm$0.01 & 0.78$\pm$0.02 & 0.51$\pm$0.01 & 0.82$\pm$0.01 & 0.85$\pm$0.01 & 0.61$\pm$0.02 & \textbf{0.90$\pm$0.01} & 0.89$\pm$0.02 \\ \midrule
\multicolumn{1}{c|}{\multirow{2}{*}{\begin{tabular}[c]{@{}c@{}}AlexNet \\ DermNet\end{tabular}}} & TPR & 0.27$\pm$0.12 & 9.53$\pm$0.54 & 0.13$\pm$0.03 & 22.8$\pm$1.13 & 41$\pm$0.48 & 5.6$\pm$0.05 & 48.23$\pm$0.87 & \textbf{62.27$\pm$0.23} \\
\multicolumn{1}{c|}{} & AUC & 0.68$\pm$0.01 & 0.74$\pm$0.01 & 0.50$\pm$0.01 & \textbf{0.94$\pm$0.01} & 0.85$\pm$0.01 & 0.89$\pm$0.01 & 0.91$\pm$0.02 & 0.87$\pm$0.02 \\ \midrule
\multicolumn{1}{c|}{\multirow{2}{*}{\begin{tabular}[c]{@{}c@{}}ResNet18 \\ CIFAR100\end{tabular}}} & TPR & 0.36$\pm$0.11 & 5.48$\pm$0.27 & 0.26$\pm$0.06 & 16.82$\pm$2.12 & 44.02$\pm$1.58 & 15.06$\pm$1.78 & 57.36$\pm$2.12 & \textbf{68.74$\pm$1.84} \\
\multicolumn{1}{c|}{} & AUC & 0.67$\pm$0.01 & 0.80$\pm$0.02 & 0.55$\pm$0.01 & 0.73$\pm$0.01 & 0.85$\pm$0.02 & 0.65$\pm$0.01 & 0.84$\pm$0.01 & \textbf{0.89$\pm$0.01} \\ \midrule
\multicolumn{1}{c|}{\multirow{2}{*}{\begin{tabular}[c]{@{}c@{}}ResNet18 \\ DermNet\end{tabular}}} & TPR & 0.27$\pm$0.11 & 0.73$\pm$0.21 & 0.06$\pm$0.01 & 32.2$\pm$0.89 & 19.93$\pm$1.23 & 17.93$\pm$2.12 & \textbf{35.6$\pm$0.96} & 31.8$\pm$0.88 \\
\multicolumn{1}{c|}{} & AUC & 0.51$\pm$0.01 & 0.59$\pm$0.01 & 0.48$\pm$0.01 & 0.52$\pm$0.01 & 0.63$\pm$0.02 & \textbf{0.66$\pm$0.01} & 0.64$\pm$0.01 & 0.62$\pm$0.01 \\ \midrule
\multicolumn{1}{c|}{\multirow{2}{*}{\begin{tabular}[c]{@{}c@{}}Diffusion Model \\ Tiny-ImageNet\end{tabular}}} & TPR & 1.10$\pm$0.50 & 2.40$\pm$0.60 & 0.25$\pm$0.02 & 1.5$\pm$0.20 & 1.80$\pm$0.30 & 1.20$\pm$0.40 & 3.20$\pm$0.30 & \textbf{4.50$\pm$0.20} \\
\multicolumn{1}{c|}{} & AUC & 0.51$\pm$0.02 & 0.54$\pm$0.01 & 0.49$\pm$0.01 & 0.54$\pm$0.01 & 0.53$\pm$0.01 & 0.51$\pm$0.01 & 0.58$\pm$0.01 & \textbf{0.59$\pm$0.01} \\ \midrule
\multicolumn{1}{c|}{\multirow{2}{*}{\begin{tabular}[c]{@{}c@{}}Diffusion Model \\ CIFAR100\end{tabular}}} & TPR & 0.80$\pm$0.10 & 1.20$\pm$0.20 & 0.11$\pm$0.01 & 1.30$\pm$0.10 & 1.80$\pm$0.10 & 1.7$\pm$0.20 & 2.1$\pm$0.20 & \textbf{3.0$\pm$0.20} \\
\multicolumn{1}{c|}{} & AUC & 0.48$\pm$0.01 & 0.61$\pm$0.01 & 0.49$\pm$0.01 & 0.47$\pm$0.01 & 0.59$\pm$0.01 & 0.52$\pm$0.01 & 0.59$\pm$0.01 & \textbf{0.62$\pm$0.01} \\ \bottomrule
\end{tabular}
\vspace{-5pt}
\end{table*}

% \vspace{-8pt}
\section{Experimental Result} \label{sec:exp}
This section presents the empirical analysis of the proposed FedMIA framework in terms of experimental setting, attack effectiveness, robustness.
\subsection{Experimental Setup} \label{subsec: MIA-setup}
\noindent\textbf{Dataset \& Models.} We employed three image datasets: CIFAR-100 \citep{cifardataset}, DermNet \citep{dernment}, and Tiny-ImageNet \citep{le2015tiny}. Additionally, we implemented three models: ResNet18 \cite{he2016deep} and AlexNet \cite{krizhevsky2012imagenet} for the classification task, and the Latent Diffusion Model \cite{rombach2022high} for the generative task.

\noindent\textbf{Federated Setting.}
We consider horizontal federated learning with two typical structures: FedAvg \citep{mcmahan2016federated}, which transfers models or gradients in classification tasks, and FedEmbedding \citep{liang2025diffusion}, which transfers prompts in generative tasks. We evaluate scenarios with 5–30 clients in FL, up to 300 communication rounds, and the data samples of each client range from 1,000 to 10,000. Local training involves 1 to 9 epochs, and we explore different Non-IID extents using the Dirichlet distribution \( dir(\beta) \), with values of \( \beta = 0.1, 1, 10, \infty \) (IID). If there are no additional instructions, each experiment has 10 clients, 300 synchronous communication rounds.

\noindent\textbf{MIAs.} We conducted a comprehensive comparison of our methods, FedMIA-I and FedMIA-II, against six baseline attack methods: Blackbox-Loss \cite{yeom2018privacy}, Grad-Cosine \cite{li2022effective}, Grad-Norm \cite{nasr2018machine}, Loss-Series \cite{gu2022cs}, Avg-Cosine \cite{li2022effective}, and Grad-Diff \cite{li2022effective}. FedMIA-I is utilizes the model loss measurement \cite{yeom2018privacy}, while FedMIA-II employs the Grad-Cosine measurement \cite{li2022effective}, as described in Eq. \eqref{eq:MDM}.

% and we classified existing methods into three types, i.e. \textbf{MIA I}, \textbf{MIA II} and \textbf{MIA III}, based on whether the attacks utilize temporal information and spatial information. \textbf{MIA I} corresponds to the method that only uses single temporal information and single spatial information, including the following attacks: black-box attack \citep{yeom2018privacy} (referred to as Loss-I), grad-norm attack \citep{nasr2019comprehensive} (Grad-Norm), gradient-diff attack \citep{li2022effective}, and Cos attack using single round and single client information (Cos-I).
% \textbf{MIA II} corresponds to the method that uses multiple temporal information and single spatial information, including the following attacks: fed-loss attack \citep{li2022effective} (Loss-II) and Cos attack based on multi-round synchronization (Cos-II). \textbf{MIA III} corresponds to the method that uses multiple temporal information and multiple spatial information, including Loss-III and Cos-III.
\noindent\textbf{Defenses.}
We evaluate the robustness of FedMIA against six defense methods, including Gradient Perturbation (Perturb) \citep{geyer2017differentially, zheng2021federated}, Gradient Sparsification (Sparse) \citep{gupta2018distributed}, MixUp \citep{zhang2017mixup, gu2023fedpass}, Data Augmentation \citep{shorten2019survey}, Data Sampling \citep{li2021sample}, and a combination of Data Augmentation + Sampling.

\noindent\textbf{Evaluation metric.} We use the metrics AUC and TPR@FPR \citep{carlini2022membership} to specifically assess the leakage of the most vulnerable samples to attacks, where TPR@FPR refers to the True Positive Rate (TPR) at a specific False Positive Rate (FPR, a.k.a. Type-I Error Rate) in binary classification. Specifically, we pay particular attention to the TPR when the FPR is very low, such as FPR values of 0.1\%.
Moreover, we test the attack effectiveness against different defense methods. Since the attack effectiveness depends on the parameters of defenses, e.g., for gradient perturbation, if more noise is added, the attack effectiveness becomes weaker, but the model utility is largely influenced. Therefore, we consider the attack effectiveness under different parameters of each defense method and obtain the tradeoff between utility loss (the test error rate) and attack effectiveness (TPR@FPR = 0.1\%). Furthermore, the effectiveness of the attack can be measured using \textbf{hypervolume (HV)} \cite{zitzler2004indicator}, which, in our case, refers to the area between the Pareto frontiers and the unit box. A larger hypervolume indicates a better privacy-utility trade-off, representing that the attack is less effective.

More experimental setup details are left in Appendix A.

\subsection{FedMIA vs Others} \label{subsec:exp-attack}
The comparison results of all attacks are presented in the Tab. \ref{tab:MIA}. We can draw the following three conclusions:
\begin{itemize}
    \item FedMIA generally outperforms other MIA methods across all experiments, as indicated by higher TPR@FPR=0.1\% and AUC metrics. For example, FedMIA-II achieves a TPR@FPR=0.1\% of (66.98 $\pm$ 1.74)\%, which is significantly higher than the next best method with (54.66 $\pm$ 1.22)\% on AlexNet-CIFAR100.
    \item Blackbox-Loss \cite{yeom2018privacy} and Grad-Cosine \cite{li2022effective} show relatively low TPR and AUC values, significantly lagging behind FedMIA. For example, in the AlexNet DermNet task, Blackbox-Loss and Grad-Cosine both result in low TPRs (around 0.27\%), while FedMIA achieves a much higher TPR of (62.27 $\pm$ 0.23)\%.
    \item FedMIA performs well in both  classification tasks and generative task. Specifically, FedMIA also leads, but the performance gap between FedMIA and other methods like Blackbox-Loss or Grad-Cosine is more pronounced. For example, FedMIA achieves a TPR of (3.0 $\pm$ 0.20)\% and an AUC of (0.62 $\pm$ 0.01)\%, while Blackbox-Loss and Grad-Cosine have much lower TPR and AUC values (around (1.7 $\pm$ 0.20)\% and (0.52 $\pm$ 0.01)\%, respectively) in  CIFAR100 with diffusion model.
\end{itemize}

\begin{figure}[ht]
    \flushleft 
   \centering
    \subfigure[Original training images]{
        \begin{minipage}[t]{1.0\linewidth}
        \centering
        \includegraphics[width=3.4in]{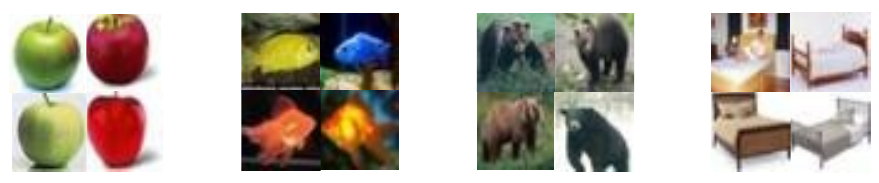}
        %\caption{fig1}
        \end{minipage}%
    }%
    
    \vspace{-8pt}
    \subfigure[Generated images]{
        \begin{minipage}[t]{1.0\linewidth}
        \centering
        \includegraphics[width=3.4in]{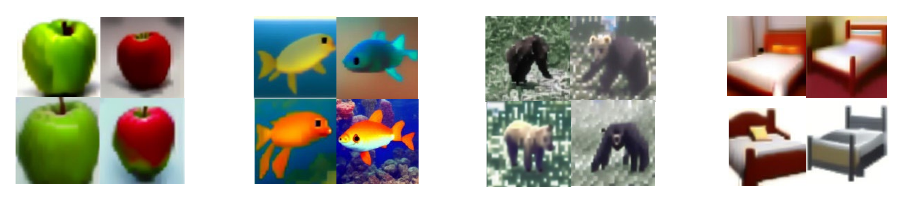}
        %\caption{fig1}
        \end{minipage}%
    }%
    \vspace{-8pt}
    \caption{Original training images  and generated images based on uploaded embeddings via latent diffusion model.}
    % \vspace{-5pt}
    \label{Fig:Gen}
\end{figure}

Furthermore, we present the generated images based on the embeddings alongside the original images, which are considered as members by the proposed FedMIA-II, as shown in Fig. \ref{Fig:Gen}. The results indicate that the two types of images are highly similar, demonstrating that the target data identified by FedMIA as a member is effectively trained on the corresponding embedding. This observation highlights the effectiveness of the FedMIA attack.

\begin{table*}[htbp]
\centering
\caption{The hypervolume of various attack methods under defense strategies with AlexNet and ResNet on CIFAR100. The smaller Hypervolume indicates the better attack effectiveness.}
\vspace{4pt}
\begin{tabular}{@{}cccccccc@{}}
\toprule
 &                & \begin{tabular}[c]{@{}c@{}} Perturb \\ \cite{zheng2021federated}  \end{tabular}           & \begin{tabular}[c]{@{}c@{}} Sparse  \\  \cite{gupta2018distributed} \end{tabular}          & \begin{tabular}[c]{@{}c@{}} Mixup \\ \citep{zhang2017mixup}       \end{tabular}      & \begin{tabular}[c]{@{}c@{}} Sampling \\ \citep{li2021sample}  \end{tabular}         & \begin{tabular}[c]{@{}c@{}} Data Aug \\ \citep{shorten2019survey}   \end{tabular}        &  \begin{tabular}[c]{@{}c@{}} Data Aug  \\ + Sampling    \end{tabular}  \\ \cmidrule(l){1-8} 
                        \multirow{6}{*}{\begin{tabular}[c]{@{}c@{}} AlexNet \\ CIFAR100 \end{tabular}} & Blackbox-Loss \cite{yeom2018privacy}  & 0.3543          & 0.3533          & 0.3328          & 0.3491          & 0.3395          & 0.3422          \\
                         & Loss-Series \cite{gu2022cs}    & 0.3252          & 0.3377          & 0.3554          & 0.3464          & 0.3375          & 0.3427          \\
                         & Grad-Cosine \cite{li2022effective}    & 0.304           & 0.3287          & 0.2845          & 0.3365          & 0.3247          & 0.3379          \\
                         & Avg-Cosine \cite{li2022effective}& 0.3421          & 0.3421          & 0.3334          & 0.3407          & 0.3248          & 0.3376          \\
                         & FedMIA-I (ours)       & 0.3611          & 0.3593          & 0.3609          & 0.3373          & 0.3202          & \textbf{0.3359} \\
                         & FedMIA-II (ours)    & \textbf{0.2702} & \textbf{0.3085} & \textbf{0.2588} & \textbf{0.3325} & \textbf{0.3175} & 0.3361          \\ \midrule
\multirow{6}{*}{\begin{tabular}[c]{@{}c@{}} ResNet \\ CIFAR100 \end{tabular}}  
                         & Blackbox-Loss \cite{yeom2018privacy}  & 0.4338          & 0.4307          & 0.4538          & 0.4568          & 0.5833          & 0.5815          \\
                         & Loss-Series \cite{gu2022cs}    & 0.4126          & 0.4092          & 0.458           & 0.4555          & 0.5823          & 0.5821          \\
                         & Grad-Cosine \cite{li2022effective}    & 0.3261          & 0.3545          & 0.3905          & 0.4407          & 0.5751          & 0.5779          \\
                         & Avg-Cosine \cite{li2022effective} & 0.4064          & 0.4059          & 0.4533          & 0.4537          & 0.571           & 0.5745          \\
                         & FedMIA-I (ours)      & 0.4438          & 0.4392          & 0.4736          & 0.4446          & \textbf{0.5665} & 0.574           \\
                         & FedMIA-II (ours)    & \textbf{0.2969} & \textbf{0.3004} & \textbf{0.4302} & \textbf{0.4425} & 0.5693          & \textbf{0.5739} \\ \bottomrule
\end{tabular}
\label{tab:defenses}
\end{table*}

\begin{figure*}[t]
    \centering
    
        \subfigure[IID]{
        \includegraphics[scale=0.26]{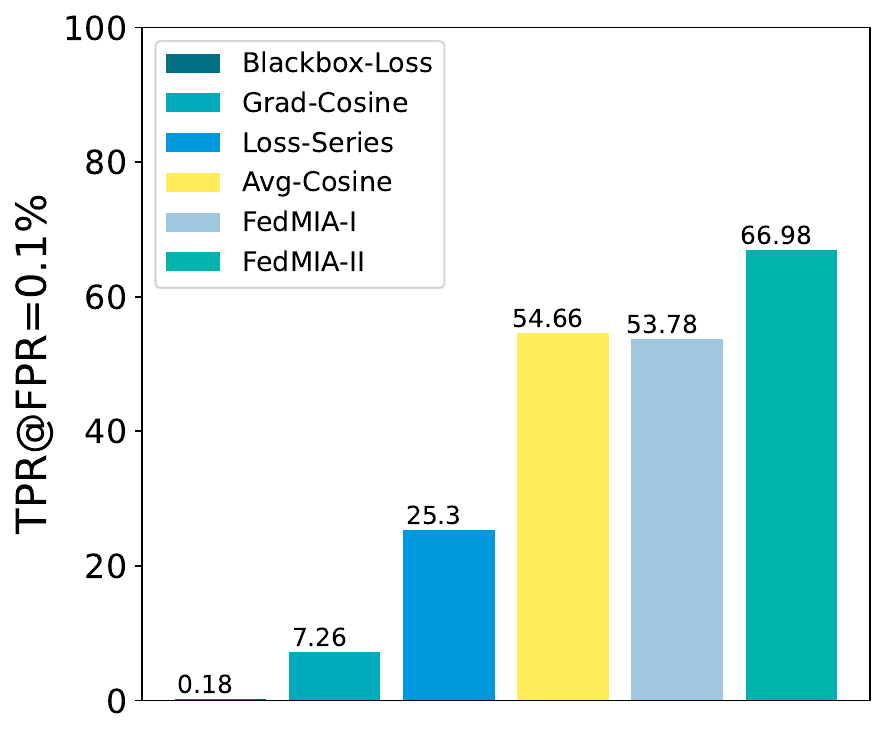} 
    }
    \subfigure[$\beta = 10$]{
        \includegraphics[scale=0.26]{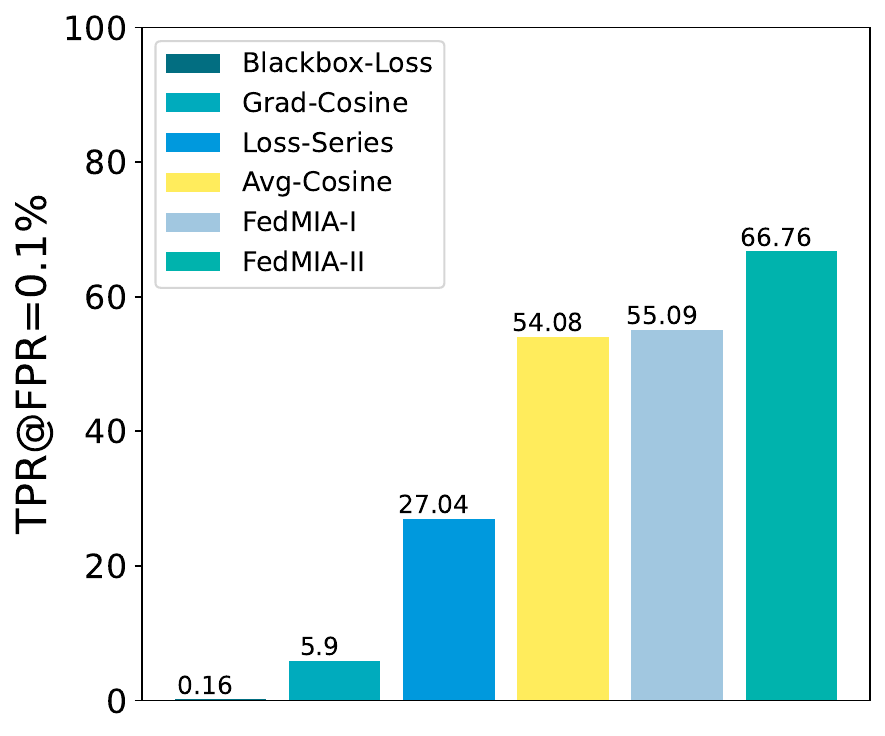} 
    }
    \subfigure[$\beta = 1$]{
        \includegraphics[scale=0.26]{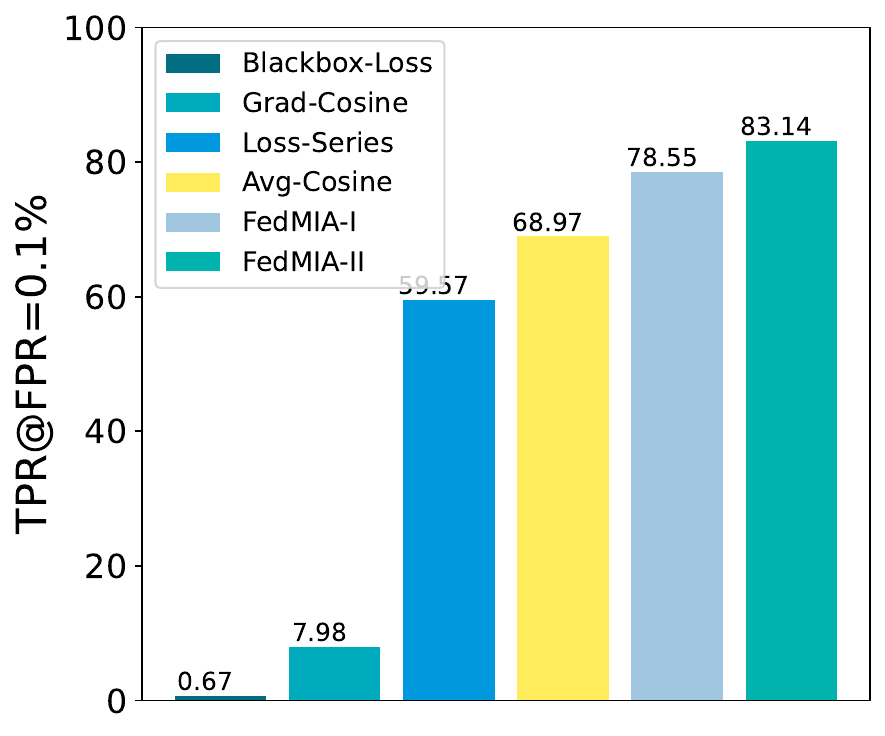} 
    }
    \subfigure[$\beta = 0.1$]{
        \includegraphics[scale=0.26]{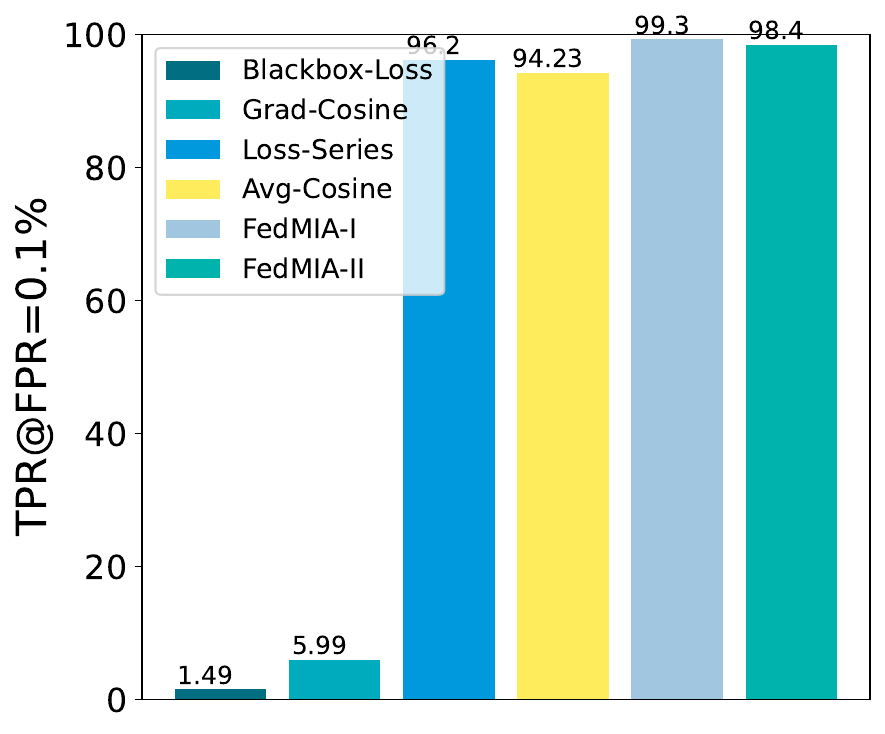} 
    }

    \subfigure[IID]{
        \includegraphics[scale=0.26]{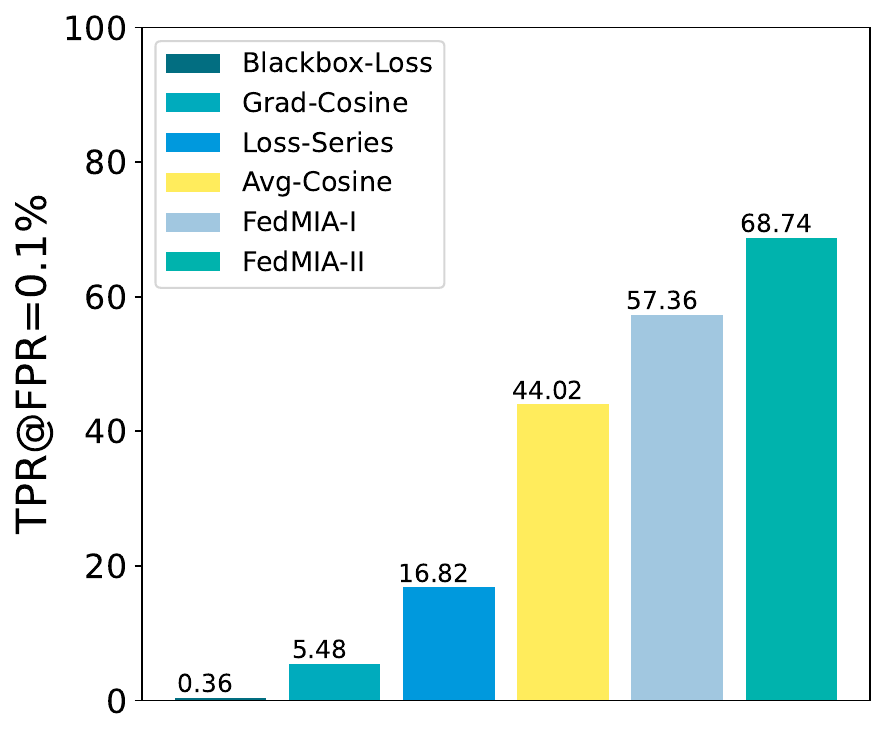} 
    }
    \subfigure[$\beta = 10$]{
        \includegraphics[scale=0.26]{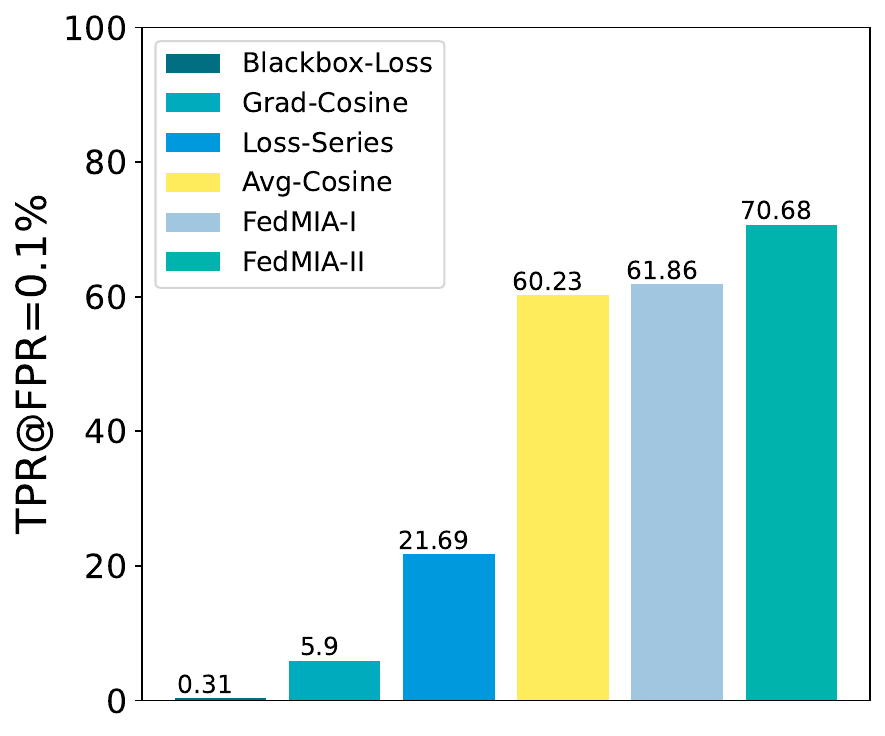} 
    }
    \subfigure[$\beta = 1$]{
        \includegraphics[scale=0.26]{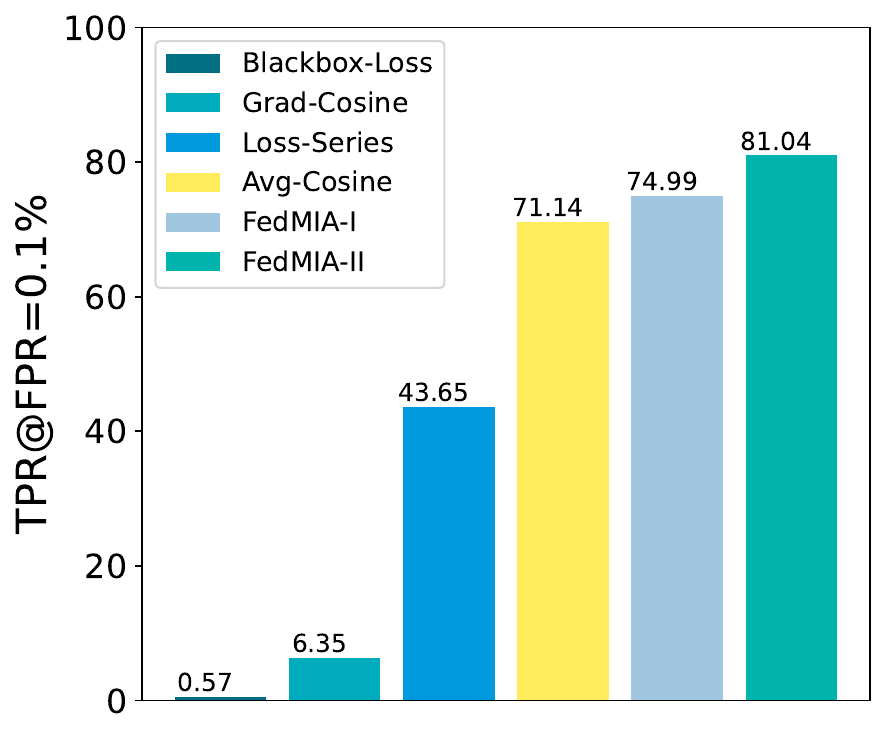} 
    }
    \subfigure[$\beta = 0.1$]{
        \includegraphics[scale=0.26]{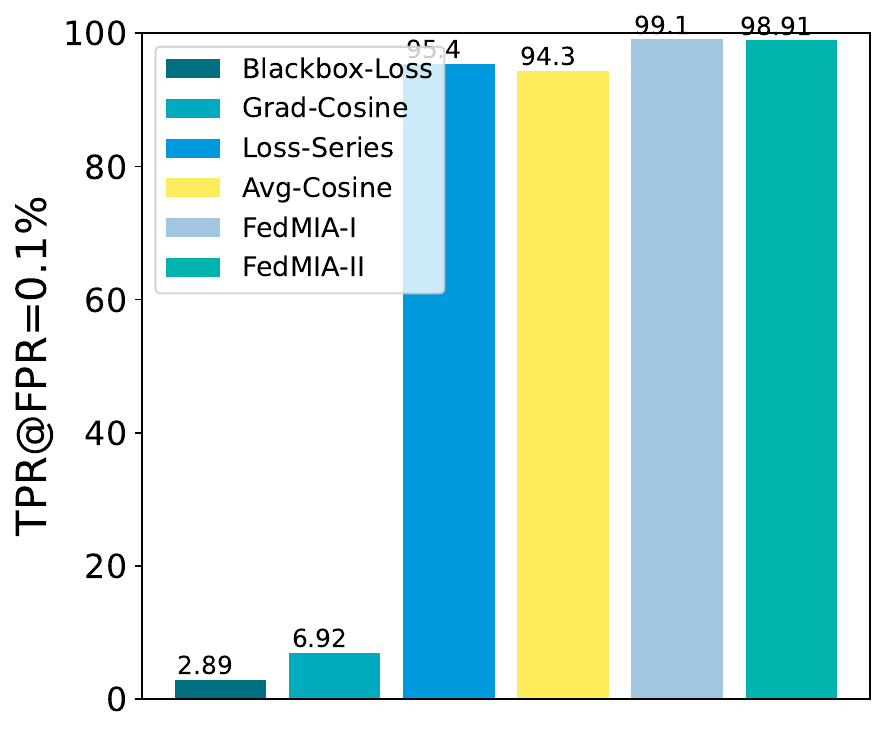} 
    }
    
    \DeclareGraphicsExtensions.
    \vspace{-8pt}
    \caption{This set of figures shows the attack effects (TPR@FPR=0.1\%) of various attacks (Blackbox-Loss\cite{yeom2018privacy}, Grad-Cosine\cite{li2022effective}, Loss-Series\cite{gu2022cs}, Avg-Cosine\cite{li2022effective}, FedMIA-I and FedMIA-II) on AlexNet and ResNet18 (the first and second row respectively) under IID and three Non-IID settings. }
    \label{Fig:non_iid results}
    \vspace{-10pt}
\end{figure*}

\subsection{Robustness}\label{subsec:robustness}
This section illustrates the robustness of FedMIA against six defense methods, varying degrees of Non-IID, different client counts, communication rounds, and local epochs.

\noindent\textbf{FedMIA against different defense methods.}
Tab. \ref{tab:defenses} and Appendix B presents the hypervolume and privacy-utility tradeoff against different defense methods. We can obtain: 1)FedMIA-I (ours) consistently performs better than other MIA methods across all defense strategies for AlexNet: Example: For AlexNet-CIFAR100, under Mixup, FedMIA-I achieves a hypervolume of 0.3609, outperforming methods like Blackbox-Loss (0.3328) and Loss-Series (0.3554); 2) Combining data augmentation and data sampling combining in an appropriate manner may have the best defense effectiveness (achieves the largest HV volume); 3) Even the strongest defense with combining data augmentation and data sampling still exist privacy leakage when  preserving the model performance (see Appendix B).

% In order to analyze the relationship between FL membership leakage and certain FL setting factors, we conducted additional experiments with different Non-IID extent, communication rounds, numbers of clients, numbers of samples, and local epochs. We use Cos-III attack for all experiments ss we have shown that it demonstrates good attack performance in all scenarios. This subsection uses TPR@FPR=1\% as the indicator to evaluate the MIA effect. The results and analysis using the Area Under Curve (AUC) value as the indicator can be found in Appendix \ref{app:auc results}.
\begin{figure*}[htbp]
    \centering
    
        \subfigure[AlexNet-CIFAR100]{
        \includegraphics[scale=0.45]{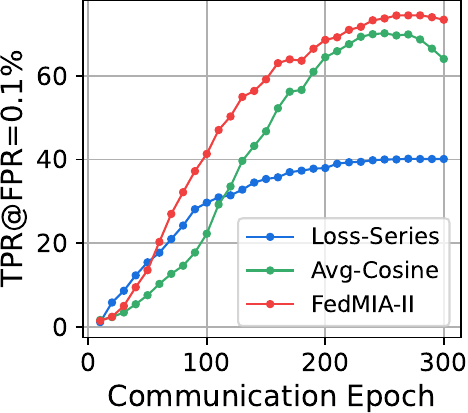} 
    }
    \subfigure[AlexNet-CIFAR100]{
        \includegraphics[scale=0.45]{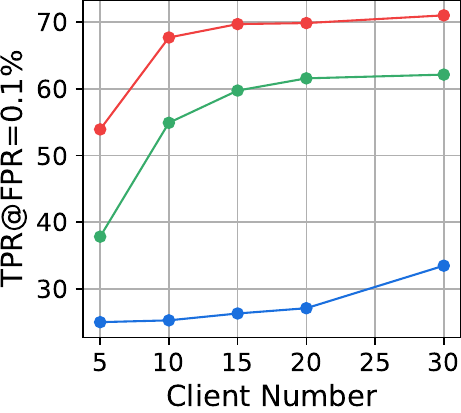} 
    }
    \subfigure[AlexNet-CIFAR100]{
        \includegraphics[scale=0.45]{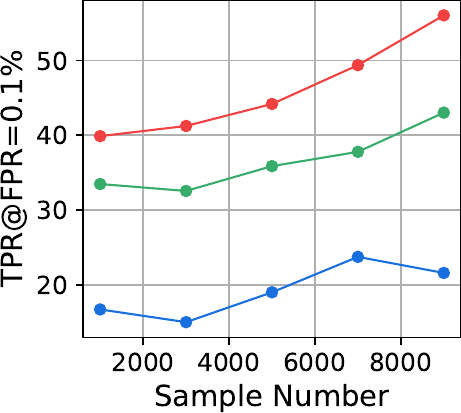} 
    }
    \subfigure[AlexNet-CIFAR100]{
        \includegraphics[scale=0.45]{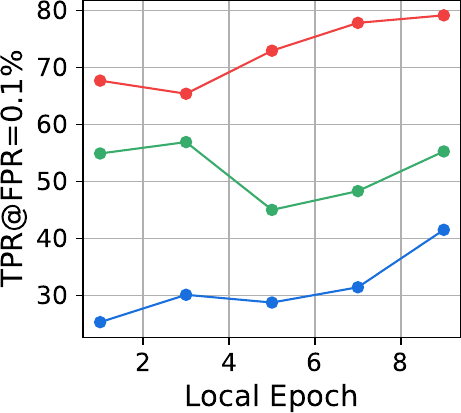} 
    }

    \subfigure[ResNet18-CIFAR100]{
        \includegraphics[scale=0.45]{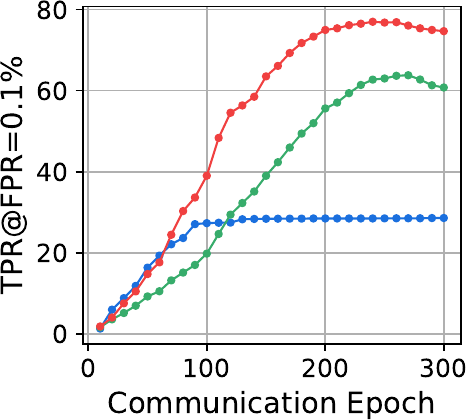} 
    }
    \subfigure[ResNet18-CIFAR100]{
        \includegraphics[scale=0.45]{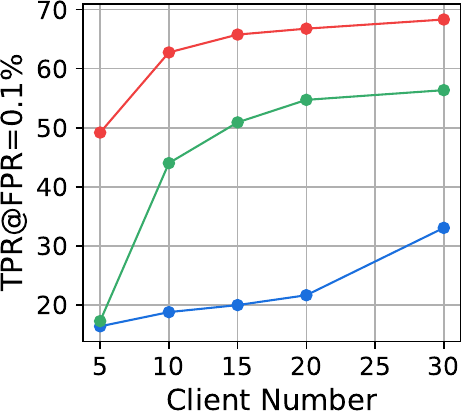} 
    }
    \subfigure[ResNet18-CIFAR100]{
        \includegraphics[scale=0.45]{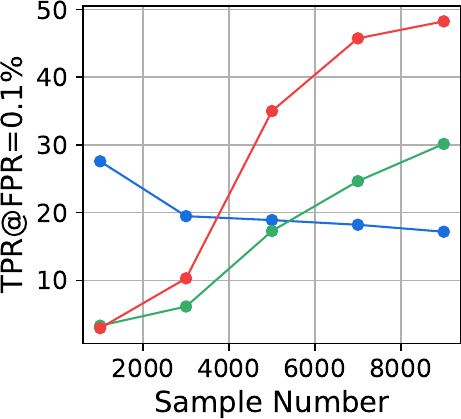} 
    }
    \subfigure[ResNet18-CIFAR100]{
        \includegraphics[scale=0.45]{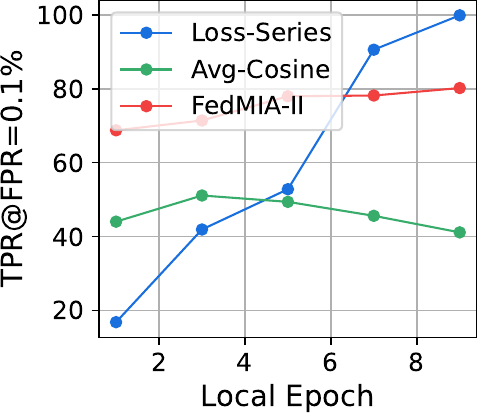} 
    }
    \vspace{-8pt}
    \DeclareGraphicsExtensions.
    \caption{This set of figures shows the attack effects (TPR@FPR=0.1\%) of various attacks (blue line: Loss-series \citep{gu2022cs}, green line: Avg-Cosine \citep{li2022effective} and red line: FedMIA-II) on AlexNet and ResNet18 (the first and second row respectively) under four settings. The four columns of the graph group show the results of different communication rounds, client numbers, data volumes and local epochs settings respectively.}
    \label{Fig:fl_settings}
    \vspace{-10pt}
\end{figure*}

\noindent\textbf{Non-IID extent.} We investigate the impact of non-IID on MIA attacks. Following \citep{hsu2019measuring}, the basic assumption of non-iid simulation in this part is that the labels of each client's training data follow the Dirichlet distribution. $\beta$ is the core parameter controlling the distribution difference and the smaller the $\beta$, the greater the degree of non-iid.  We report the performance of the FedMIA-II attack on the CIFAR-100 dataset in a table. We control the degree of non-IID by adjusting the parameter alpha, where a smaller $\beta$ indicates a more severe Non-IID condition. 

Based on Fig.\ref{Fig:non_iid results}, we can observe the following: 1) the proposed method is the strongest attacks with various Non-IID extent, e.g., the TPR@FPR =0.1\% for FedMIA-II achieves the largest value 67\% in AlexNet-CIFAR100. 2) As non-IID increases, TPR shows a increasing trend. One reason is when non-IID becomes severe, such as when a client contains only a few classes, MIA attacks themselves become easier. An extreme example is when a client contains only one class, in which case we can achieve a strong baseline by simply judging based on the sample labels.

\noindent\textbf{Communication round.} {As for the benefits of synchronous rounds to our scheme, it can be observed in Figure \ref{Fig:fl_settings}(a) and (e) that the attack effect of our scheme increases rapidly in most epochs as the synchronous communication progresses. At epoch=200, the TPR@FPR=0.1\% of Ours exceeds 0.4, which is twice as high as that of the Avg-Cosine attack and Loss-series attack. After that, the attack effect shows a slight decrease (about 5\% on TPR@FPR=0.1\%) and a similar trend can also be observed in the curve of the Avg-Cosine attack. This phenomenon may be attributed to the fact that the information obtained in the later epochs is not as helpful for the membership leakage attack as the information acquired in the previous epochs.}

\noindent\textbf{Number of Clients.} Figure \ref{Fig:fl_settings}(b) and (f) illustrate the effects of membership inference attacks on AlexNet and ResNet18, while varying the number of clients from 2 to 20. As depicted in the figures, our two attacks outperform the baselines in most cases, indicating their significantly higher effectiveness. Furthermore, the gradually rising red and blue curves indicates that as the number of clients increases, the target model becomes more vulnerable to  our MIA scheme.

\noindent\textbf{Number of Samples.} Figure \ref{Fig:fl_settings}(c) and (g) demonstrate the impact of varying the number of samples (ranging from 500 to 5000) on the attack effects of MIAs on AlexNet and ResNet18. 
Regardless of the increase in the number of samples, the attack effect of our scheme remains consistently high and even demonstrates notable improvement on AlexNet. The TPR@FPR=0.1\% of ours consistently exceeds twice that of the baseline in both subfigures. This indicates that our attack scheme maintains a significant advantage as the training data increases.

%As the number of samples increases, the attack effects of the fed loss attack and cosine attack gradually decrease. However, our attack method does not experience a significant decrease in effectiveness and even exhibits an observable upward trend for AlexNet. This suggests that increasing the number of training samples per client can provide some resistance against many previous MIAs, but it does not offer protection against our attack method.
\noindent\textbf{Local Epoch.} Figure \ref{Fig:fl_settings}(d) and (h) illustrate the effects of MIAs on AlexNet and ResNet18 as the number of local epochs varies from 1 to 9. As the number of local epochs increases, the effectiveness of our attack method and the fed loss attack significantly improve while the enhancement effect of the Avg-Cosine attack is not evident. This demonstrates that an increase in the number of local epochs may render the model more susceptible to MIA.
\section{Discussion and Conclusion}
While advantages brought by FL can be ascribed to, by and large, the principle of ``one for all and all for one", this paper shows that information shared by all clients through a semi-honest server can actually be adversely exploited to launch very effective membership inference attacks. Specifically, 
this paper introduce FedMIA, a novel Membership Inference Attack (MIA) method by leveraging updates from non-target clients and applies a one-tailed likelihood-ratio hypothesis test. This enables the inference of target data membership without requiring access to auxiliary datasets or making strong assumptions about the training process. Through extensive experiments, we demonstrated that FedMIA is highly effective across various federated learning configurations, including both classification and generative tasks, and remains robust against common defense methods, Non-IID data, and different client setups.

% One effective defense may involve data preprocessing techniques, such as local shuffling and stratification \cite{hu2023defenses}. These strategies aim to obfuscate the patterns within participants' local datasets, making it more challenging for an adversary to distinguish the presence or absence of specific data points. 

% In summary, we aim to assist researchers in gaining a better understanding of privacy leakage in federated learning and to introduce the more strong defense methods against MIA.

Conventional FL privacy defenses (perturbation, sparsification, mixup) prove ineffective against FedMIA due to attackers’ exploitation of cross-client information patterns as shown in Sect. 4.3. While secure aggregation via MPC \cite{fereidooni2021safelearn}\cite{gehlhar2023safefl}/HE \cite{wibawa2022homomorphic}\cite{zhang2020batchcrypt} blocks FedMIA by encrypting individual updates, their computational/communication costs hinder practical deployment. This exposes an urgent need for MIA defenses specifically to resist attackers from obtaining valuable information from non-target updates.

\clearpage

\clearpage
\bibliographystyle{ieeenat_fullname}
\bibliography{reference}

\begin{thebibliography}{51}
\providecommand{\natexlab}[1]{#1}
\providecommand{\url}[1]{\texttt{#1}}
\expandafter\ifx\csname urlstyle\endcsname\relax
  \providecommand{\doi}[1]{doi: #1}\else
  \providecommand{\doi}{doi: \begingroup \urlstyle{rm}\Url}\fi

\bibitem[Aboulmira et~al.(2022)Aboulmira, Hrimech, and Lachgar]{dernment}
Amina Aboulmira, Hamid Hrimech, and Mohamed Lachgar.
\newblock Comparative study of multiple cnn models for classification of 23 skin diseases.
\newblock \emph{International Journal of Online \& Biomedical Engineering}, 18\penalty0 (11), 2022.

\bibitem[Aerni et~al.(2024)Aerni, Zhang, and Tram{\`e}r]{aerni2024evaluations}
Michael Aerni, Jie Zhang, and Florian Tram{\`e}r.
\newblock Evaluations of machine learning privacy defenses are misleading.
\newblock \emph{arXiv preprint arXiv:2404.17399}, 2024.

\bibitem[Carlini et~al.(2022)Carlini, Chien, Nasr, Song, Terzis, and Tramer]{carlini2022membership}
Nicholas Carlini, Steve Chien, Milad Nasr, Shuang Song, Andreas Terzis, and Florian Tramer.
\newblock Membership inference attacks from first principles.
\newblock In \emph{2022 IEEE Symposium on Security and Privacy (SP)}, pages 1897--1914. IEEE, 2022.

\bibitem[Choo et~al.(2020)Choo, Tramer, Carlini, and Papernot]{choo2020label}
Christopher A~Choquette Choo, Florian Tramer, Nicholas Carlini, and Nicolas Papernot.
\newblock Label-only membership inference attacks.
\newblock \emph{arXiv preprint arXiv:2007.14321}, 2020.

\bibitem[Fereidooni et~al.(2021)Fereidooni, Marchal, Miettinen, Mirhoseini, M{\"o}llering, Nguyen, Rieger, Sadeghi, Schneider, Yalame, et~al.]{fereidooni2021safelearn}
Hossein Fereidooni, Samuel Marchal, Markus Miettinen, Azalia Mirhoseini, Helen M{\"o}llering, Thien~Duc Nguyen, Phillip Rieger, Ahmad-Reza Sadeghi, Thomas Schneider, Hossein Yalame, et~al.
\newblock Safelearn: Secure aggregation for private federated learning.
\newblock In \emph{2021 IEEE Security and Privacy Workshops (SPW)}, pages 56--62. IEEE, 2021.

\bibitem[Gehlhar et~al.(2023)Gehlhar, Marx, Schneider, Suresh, Wehrle, and Yalame]{gehlhar2023safefl}
Till Gehlhar, Felix Marx, Thomas Schneider, Ajith Suresh, Tobias Wehrle, and Hossein Yalame.
\newblock Safefl: Mpc-friendly framework for private and robust federated learning.
\newblock In \emph{2023 IEEE Security and Privacy Workshops (SPW)}, pages 69--76. IEEE, 2023.

\bibitem[Geiping et~al.(2020)Geiping, Bauermeister, Dr{\"o}ge, and Moeller]{geiping2020inverting}
Jonas Geiping, Hartmut Bauermeister, Hannah Dr{\"o}ge, and Michael Moeller.
\newblock Inverting gradients-how easy is it to break privacy in federated learning?
\newblock \emph{Advances in Neural Information Processing Systems}, 33:\penalty0 16937--16947, 2020.

\bibitem[Geyer et~al.(2017)Geyer, Klein, and Nabi]{geyer2017differentially}
Robin~C Geyer, Tassilo Klein, and Moin Nabi.
\newblock Differentially private federated learning: A client level perspective.
\newblock \emph{arXiv preprint arXiv:1712.07557}, 2017.

\bibitem[Gu et~al.(2023)Gu, Luo, Kang, Fan, and Yang]{gu2023fedpass}
Hanlin Gu, Jiahuan Luo, Yan Kang, Lixin Fan, and Qiang Yang.
\newblock Fedpass: Privacy-preserving vertical federated deep learning with adaptive obfuscation.
\newblock \emph{arXiv e-prints}, pages arXiv--2301, 2023.

\bibitem[Gu et~al.(2022)Gu, Bai, and Xu]{gu2022cs}
Yuhao Gu, Yuebin Bai, and Shubin Xu.
\newblock Cs-mia: Membership inference attack based on prediction confidence series in federated learning.
\newblock \emph{Journal of Information Security and Applications}, 67:\penalty0 103201, 2022.

\bibitem[Gupta and Raskar(2018)]{gupta2018distributed}
Otkrist Gupta and Ramesh Raskar.
\newblock Distributed learning of deep neural network over multiple agents.
\newblock \emph{Journal of Network and Computer Applications}, 116:\penalty0 1--8, 2018.

\bibitem[Haddadpour et~al.(2021)Haddadpour, Kamani, Mokhtari, and Mahdavi]{haddadpour2021federated}
Farzin Haddadpour, Mohammad~Mahdi Kamani, Aryan Mokhtari, and Mehrdad Mahdavi.
\newblock Federated learning with compression: Unified analysis and sharp guarantees.
\newblock In \emph{International Conference on Artificial Intelligence and Statistics}, pages 2350--2358. PMLR, 2021.

\bibitem[He et~al.(2016)He, Zhang, Ren, and Sun]{he2016deep}
Kaiming He, Xiangyu Zhang, Shaoqing Ren, and Jian Sun.
\newblock Deep residual learning for image recognition.
\newblock In \emph{Proceedings of the IEEE conference on Computer Vision and Pattern Recognition (CVPR)}, pages 770--778, 2016.

\bibitem[He et~al.(2024)He, Xu, Zhang, Xu, and Yan]{he2024enhance}
Xinlong He, Yang Xu, Sicong Zhang, Weida Xu, and Jiale Yan.
\newblock Enhance membership inference attacks in federated learning.
\newblock \emph{Computers \& Security}, 136:\penalty0 103535, 2024.

\bibitem[Hsu et~al.(2019)Hsu, Qi, and Brown]{hsu2019measuring}
Tzu-Ming~Harry Hsu, Hang Qi, and Matthew Brown.
\newblock Measuring the effects of non-identical data distribution for federated visual classification.
\newblock \emph{arXiv preprint arXiv:1909.06335}, 2019.

\bibitem[Hu et~al.(2023)Hu, Zhang, Salcic, Sun, Choo, and Dobbie]{hu2023source}
Hongsheng Hu, Xuyun Zhang, Zoran Salcic, Lichao Sun, Kim-Kwang~Raymond Choo, and Gillian Dobbie.
\newblock Source inference attacks: Beyond membership inference attacks in federated learning.
\newblock \emph{IEEE Transactions on Dependable and Secure Computing}, 2023.

\bibitem[Hui et~al.(2021)Hui, Yang, Yuan, Burlina, Gong, and Cao]{hui2021practical}
Bo Hui, Yuchen Yang, Haolin Yuan, Philippe Burlina, Neil~Zhenqiang Gong, and Yinzhi Cao.
\newblock Practical blind membership inference attack via differential comparisons.
\newblock \emph{arXiv preprint arXiv:2101.01341}, 2021.

\bibitem[Kang et~al.(2023)Kang, Gu, Tang, He, Zhang, He, Han, Fan, and Yang]{kang2023optimizing}
Yan Kang, Hanlin Gu, Xingxing Tang, Yuanqin He, Yuzhu Zhang, Jinnan He, Yuxing Han, Lixin Fan, and Qiang Yang.
\newblock Optimizing privacy, utility and efficiency in constrained multi-objective federated learning.
\newblock \emph{arXiv preprint arXiv:2305.00312}, 2023.

\bibitem[Kone{\v{c}}n{\`y} et~al.(2016)Kone{\v{c}}n{\`y}, McMahan, Ramage, and Richt{\'a}rik]{konevcny2016federated}
Jakub Kone{\v{c}}n{\`y}, H~Brendan McMahan, Daniel Ramage, and Peter Richt{\'a}rik.
\newblock Federated optimization: Distributed machine learning for on-device intelligence.
\newblock \emph{arXiv preprint arXiv:1610.02527}, 2016.

\bibitem[Krizhevsky et~al.()Krizhevsky, Nair, and Hinton]{cifardataset}
Alex Krizhevsky, Vinod Nair, and Geoffrey Hinton.
\newblock Cifar-10 (canadian institute for advanced research).

\bibitem[Krizhevsky et~al.(2012)Krizhevsky, Sutskever, and Hinton]{krizhevsky2012imagenet}
Alex Krizhevsky, Ilya Sutskever, and Geoffrey~E Hinton.
\newblock Imagenet classification with deep convolutional neural networks.
\newblock \emph{Advances in neural information processing systems}, 25, 2012.

\bibitem[Le and Yang(2015)]{le2015tiny}
Ya Le and Xuan Yang.
\newblock Tiny imagenet visual recognition challenge.
\newblock \emph{CS 231N}, 7\penalty0 (7):\penalty0 3, 2015.

\bibitem[Li et~al.(2021)Li, Zhang, Tan, Qin, Wang, and Li]{li2021sample}
Anran Li, Lan Zhang, Juntao Tan, Yaxuan Qin, Junhao Wang, and Xiang-Yang Li.
\newblock Sample-level data selection for federated learning.
\newblock In \emph{IEEE INFOCOM 2021-IEEE Conference on Computer Communications}, pages 1--10. IEEE, 2021.

\bibitem[Li et~al.(2022)Li, Li, and Ribeiro]{li2022effective}
Jiacheng Li, Ninghui Li, and Bruno Ribeiro.
\newblock Effective passive membership inference attacks in federated learning against overparameterized models.
\newblock In \emph{The Eleventh International Conference on Learning Representations}, 2022.

\bibitem[Liang et~al.(2025)Liang, Zhong, Gu, Lu, Tang, Dai, Huang, Fan, and Yang]{liang2025diffusion}
Jinglin Liang, Jin Zhong, Hanlin Gu, Zhongqi Lu, Xingxing Tang, Gang Dai, Shuangping Huang, Lixin Fan, and Qiang Yang.
\newblock Diffusion-driven data replay: A novel approach to combat forgetting in federated class continual learning.
\newblock In \emph{European Conference on Computer Vision}, pages 303--319. Springer, 2025.

\bibitem[McMahan et~al.(2017)McMahan, Moore, Ramage, Hampson, and y~Arcas]{mcmahan2017communication}
Brendan McMahan, Eider Moore, Daniel Ramage, Seth Hampson, and Blaise~Aguera y Arcas.
\newblock Communication-efficient learning of deep networks from decentralized data.
\newblock In \emph{Proceedings of Artificial Intelligence and Statistics (AISTATS)}, pages 1273--1282, 2017.

\bibitem[McMahan et~al.(2016)McMahan, Moore, Ramage, and y~Arcas]{mcmahan2016federated}
H~Brendan McMahan, Eider Moore, Daniel Ramage, and Blaise~Ag{\"u}era y Arcas.
\newblock Federated learning of deep networks using model averaging.
\newblock \emph{arXiv preprint arXiv:1602.05629}, 2016.

\bibitem[Nasr et~al.(2018)Nasr, Shokri, and Houmansadr]{nasr2018machine}
Milad Nasr, Reza Shokri, and Amir Houmansadr.
\newblock Machine learning with membership privacy using adversarial regularization.
\newblock In \emph{Proceedings of the 2018 ACM SIGSAC Conference on Computer and Communications Security (CCS)}, 2018.

\bibitem[Nasr et~al.(2019)Nasr, Shokri, and Houmansadr]{nasr2019comprehensive}
Milad Nasr, Reza Shokri, and Amir Houmansadr.
\newblock Comprehensive privacy analysis of deep learning: Passive and active white-box inference attacks against centralized and federated learning.
\newblock In \emph{2019 IEEE symposium on security and privacy (SP)}, pages 739--753. IEEE, 2019.

\bibitem[Reisizadeh et~al.(2020)Reisizadeh, Mokhtari, Hassani, Jadbabaie, and Pedarsani]{reisizadeh2020fedpaq}
Amirhossein Reisizadeh, Aryan Mokhtari, Hamed Hassani, Ali Jadbabaie, and Ramtin Pedarsani.
\newblock Fedpaq: A communication-efficient federated learning method with periodic averaging and quantization.
\newblock In \emph{International Conference on Artificial Intelligence and Statistics}, pages 2021--2031. PMLR, 2020.

\bibitem[Rezaei and Liu(2020)]{rezaei2020towards}
Shahbaz Rezaei and Xin Liu.
\newblock Towards the infeasibility of membership inference on deep models.
\newblock \emph{arXiv preprint arXiv:2005.13702}, 2020.

\bibitem[Rombach et~al.(2022)Rombach, Blattmann, Lorenz, Esser, and Ommer]{rombach2022high}
Robin Rombach, Andreas Blattmann, Dominik Lorenz, Patrick Esser, and Bj{\"o}rn Ommer.
\newblock High-resolution image synthesis with latent diffusion models.
\newblock In \emph{Proceedings of the IEEE/CVF conference on computer vision and pattern recognition}, pages 10684--10695, 2022.

\bibitem[Sablayrolles et~al.(2019)Sablayrolles, Douze, Ollivier, Schmid, and J{\'e}gou]{sablayrolles2019white}
Alexandre Sablayrolles, Matthijs Douze, Yann Ollivier, Cordelia Schmid, and Herv{\'e} J{\'e}gou.
\newblock White-box vs black-box: Bayes optimal strategies for membership inference.
\newblock In \emph{International Conference on Machine Learning (ICML)}. PMLR, 2019.

\bibitem[Salem et~al.(2019)Salem, Zhang, Humbert, Fritz, and Backes]{ndss19salem}
Ahmed Salem, Yang Zhang, Mathias Humbert, Mario Fritz, and Michael Backes.
\newblock Ml-leaks: Model and data independent membership inference attacks and defenses on machine learning models.
\newblock In \emph{Annual Network and Distributed System Security Symposium (NDSS)}, 2019.

\bibitem[Shokri and Shmatikov(2015)]{shokri2015privacy}
Reza Shokri and Vitaly Shmatikov.
\newblock Privacy-preserving deep learning.
\newblock In \emph{Proceedings of the 22nd ACM SIGSAC conference on computer and communications security}, pages 1310--1321, 2015.

\bibitem[Shokri et~al.(2017)Shokri, Stronati, Song, and Shmatikov]{shokri2017membership}
Reza Shokri, Marco Stronati, Congzheng Song, and Vitaly Shmatikov.
\newblock Membership inference attacks against machine learning models.
\newblock In \emph{2017 IEEE symposium on security and privacy (SP)}, pages 3--18. IEEE, 2017.

\bibitem[Shorten and Khoshgoftaar(2019)]{shorten2019survey}
Connor Shorten and Taghi~M Khoshgoftaar.
\newblock A survey on image data augmentation for deep learning.
\newblock \emph{Journal of big data}, 6\penalty0 (1):\penalty0 1--48, 2019.

\bibitem[Song and Mittal(2020)]{song2020systematic}
Liwei Song and Prateek Mittal.
\newblock Systematic evaluation of privacy risks of machine learning models.
\newblock \emph{arXiv preprint arXiv:2003.10595}, 2020.

\bibitem[Suri et~al.(2022)Suri, Kanani, Marathe, and Peterson]{suri2022subject}
Anshuman Suri, Pallika Kanani, Virendra~J Marathe, and Daniel~W Peterson.
\newblock Subject membership inference attacks in federated learning.
\newblock \emph{arXiv preprint arXiv:2206.03317}, 2022.

\bibitem[Thapa et~al.(2022)Thapa, Arachchige, Camtepe, and Sun]{thapa2022splitfed}
Chandra Thapa, Pathum Chamikara~Mahawaga Arachchige, Seyit Camtepe, and Lichao Sun.
\newblock Splitfed: When federated learning meets split learning.
\newblock In \emph{Proceedings of the AAAI Conference on Artificial Intelligence}, pages 8485--8493, 2022.

\bibitem[Truex et~al.(2019)Truex, Liu, Gursoy, Yu, and Wei]{truex2019demystifying}
Stacey Truex, Ling Liu, Mehmet~Emre Gursoy, Lei Yu, and Wenqi Wei.
\newblock Demystifying membership inference attacks in machine learning as a service.
\newblock \emph{IEEE Transactions on Services Computing}, 2019.

\bibitem[Wibawa et~al.(2022)Wibawa, Catak, Kuzlu, Sarp, and Cali]{wibawa2022homomorphic}
Febrianti Wibawa, Ferhat~Ozgur Catak, Murat Kuzlu, Salih Sarp, and Umit Cali.
\newblock Homomorphic encryption and federated learning based privacy-preserving cnn training: Covid-19 detection use-case.
\newblock In \emph{Proceedings of the 2022 European Interdisciplinary Cybersecurity Conference}, pages 85--90, 2022.

\bibitem[Yang et~al.(2019)Yang, Liu, Chen, and Tong]{yang2019federated}
Qiang Yang, Yang Liu, Tianjian Chen, and Yongxin Tong.
\newblock Federated machine learning: Concept and applications.
\newblock \emph{ACM Transactions on Intelligent Systems and Technology (TIST)}, 10\penalty0 (2):\penalty0 1--19, 2019.

\bibitem[Yeom et~al.(2018)Yeom, Giacomelli, Fredrikson, and Jha]{yeom2018privacy}
Samuel Yeom, Irene Giacomelli, Matt Fredrikson, and Somesh Jha.
\newblock Privacy risk in machine learning: Analyzing the connection to overfitting.
\newblock In \emph{2018 IEEE 31st computer security foundations symposium (CSF)}, pages 268--282. IEEE, 2018.

\bibitem[Zari et~al.(2021)Zari, Xu, and Neglia]{zari2021efficient}
Oualid Zari, Chuan Xu, and Giovanni Neglia.
\newblock Efficient passive membership inference attack in federated learning.
\newblock \emph{arXiv preprint arXiv:2111.00430}, 2021.

\bibitem[Zhang et~al.(2020{\natexlab{a}})Zhang, Li, Xia, Wang, Yan, and Liu]{zhang2020batchcrypt}
Chengliang Zhang, Suyi Li, Junzhe Xia, Wei Wang, Feng Yan, and Yang Liu.
\newblock $\{$BatchCrypt$\}$: Efficient homomorphic encryption for $\{$Cross-Silo$\}$ federated learning.
\newblock In \emph{2020 USENIX annual technical conference (USENIX ATC 20)}, pages 493--506, 2020{\natexlab{a}}.

\bibitem[Zhang et~al.(2017)Zhang, Cisse, Dauphin, and Lopez-Paz]{zhang2017mixup}
Hongyi Zhang, Moustapha Cisse, Yann~N Dauphin, and David Lopez-Paz.
\newblock mixup: Beyond empirical risk minimization.
\newblock \emph{arXiv preprint arXiv:1710.09412}, 2017.

\bibitem[Zhang et~al.(2020{\natexlab{b}})Zhang, Zhang, Chen, and Yu]{zhang2020gan}
Jingwen Zhang, Jiale Zhang, Junjun Chen, and Shui Yu.
\newblock Gan enhanced membership inference: A passive local attack in federated learning.
\newblock In \emph{ICC 2020-2020 IEEE International Conference on Communications (ICC)}, pages 1--6. IEEE, 2020{\natexlab{b}}.

\bibitem[Zheng et~al.(2021)Zheng, Chen, Long, and Su]{zheng2021federated}
Qinqing Zheng, Shuxiao Chen, Qi Long, and Weijie Su.
\newblock Federated f-differential privacy.
\newblock In \emph{International Conference on Artificial Intelligence and Statistics}, pages 2251--2259. PMLR, 2021.

\bibitem[Zhu et~al.(2019)Zhu, Liu, and Han]{zhu2019deep}
Ligeng Zhu, Zhijian Liu, and Song Han.
\newblock Deep leakage from gradients.
\newblock \emph{Advances in neural information processing systems}, 32, 2019.

\bibitem[Zitzler and K{\"u}nzli(2004)]{zitzler2004indicator}
Eckart Zitzler and Simon K{\"u}nzli.
\newblock Indicator-based selection in multiobjective search.
\newblock In \emph{International conference on parallel problem solving from nature}, pages 832--842. Springer, 2004.

\end{thebibliography}
\clearpage
\appendix
\newpage
\section{Appendix}

% \subsection{}

\subsection{Dataset and Training Details}
\label{app:dataset}
The CIFAR-100 dataset \cite{cifardataset} consists of 100 categories with 60,000 32 $\times$ 32 color images, where 50,000 images are allocated for training and 10,000 images for testing. The Dermnet dataset \cite{dernment} includes 23 categories with a total of 19,500 images, where 15,500 images are allocated for training and 4,000 images for testing. Since the images have varying sizes, we cropped them to a size of 64$\times$64 pixels. The Tiny Imagenet datasets \cite{le2015tiny} has 100000 images of 200 classes. Each class has 500 training images, 50 validation images, and 50 test images. 
The member dataset we use is the splited training datatset of target client and the non-member datatset is consist of the one-tenth hold-out test dataset and the sum of one-tenth training datset of the other clients. In the IID setting, we uniformly and randomly distribute the samples of each class to each client. In the Non-IID setting, we make the labels of each client’s training data follow the Dirichlet distribution\cite{hsu2019measuring}.
For image generative task with diffusion model, we use 10 classes for training model and attacking membership privacy.
We run image classification tasks on AlexNet and ResNet18 with NVIDIA 2060 GPU and run image generation task on laten diffusion model with NVIDIA A100 GPU.
The training parameters details and dataset splited method of federated learning are shown in Table \ref{tab:appd-training}.

\begin{table*}[htbp]
\caption{Training parameters for federated learning in this paper}
\label{tab:appd-training}
\renewcommand\arraystretch{1.1}
\small
\centering
\begin{tabular}{@{}c|c|c|c@{}}
\toprule
Dataset                          & CIFAR100                         & Dermnet   & Tiny ImageNet                       \\ \midrule
Models                           & AlexNet, ResNet18                & AlexNet, ResNet18   & Laten Diffusion Model              \\
Communication epoch              & 300                              & 300     &20                         \\
Optimizer                        & SGD                              & SGD       & Adam                       \\
Initial learning rate            & 0.1                              & 0.1    &  0.001                         \\
Learning rate decay              &  0.99 at each epoch & 0.99 at each epoch  & Adaptive \\
Number of clients                & 10                               & 10     & 10                          \\
Training set size for one client & 5000                             & 1500   & 1000                          \\ 
Testing set size                 & 10000                            & 4500   & 1000                          \\ \bottomrule
\end{tabular}
\end{table*}

\subsection{Defense Methods}
\subsubsection{Gradient Perturbation} \label{sec:defense}
\noindent\textbf{Client-level Differential Privacy.} Differential Privacy (DP) 
\citep{geyer2017differentially,zheng2021federated} hides the membership of individual data by clipping the gradients at the client level and adding Gaussian noise. The magnitude of the noise controls the strength of privacy protection: the larger the noise, the better the privacy protection, but the worse the model's performance. In the experiment, we set the DP noise standard deviation from 0.01 to 0.5 to achieve different levels of defense.

\noindent\textbf{Gradient Quantization.} Gradient quantization \citep{reisizadeh2020fedpaq,haddadpour2021federated} is a technique used to reduce the precision of gradient updates and mitigate information leakage. This algorithm quantizes the values of gradients into discrete approximations, reducing the precision of the gradients. By reducing the detailed information in the gradients, it lowers the sensitivity to individual data and improves privacy protection. The number of bits used for quantization affects the privacy protection effectiveness, where fewer bits introduce larger gradient errors but provide better privacy protection. In the experiment, we set the number of bits from 1 to 10 to achieve different levels of defense.

\noindent\textbf{Gradient Sparsification.} The gradient sparsification algorithm \citep{gupta2018distributed,shokri2015privacy,thapa2022splitfed} reduces the risk of information leakage by setting smaller absolute value elements in the gradient to zero. The fewer non-zero elements in the gradient, the less privacy leakage occurs.
In the experiment, we set the rate of gradient elements sparsified from 0.1 to 0.99 to achieve different levels of defense.

\subsubsection{Data Replacement}
\noindent\textbf{MixUp.} MixUp method \citep{zhang2017mixup,gu2023fedpass} trains neural networks on composite
images created via linear combination of image pairs. It has been shown to improve the generalization of the neural network and stabilizes the training. The coefficient of the linear combination is sampled from a Beta Distribution. We set the Beta Distribution parameter from 1e-5 to 1e5 to achieve different levels of defense.

\noindent\textbf{Data Augmentation.} Data Augmentation \cite{shorten2019survey}  includes cropping, shifting, rotating, flipping, shearing, and color jittering. We combine these augmentation schemes in different amounts to achieve different levels of defense.

\noindent\textbf{Data Sampling.} In each local training epoch, clients may choose to sample a portion of the training data instead of using the entire dataset \cite{li2021sample}. We set the portion from 0.1 to 1.0 to achieve different levels of defense.

\begin{figure*}[ht]
    \centering
    \tiny
    \hspace{-0.2cm}
    \subfigure[\tiny AlexNet-CIFAR100 Blackbox-Loss]{
        \includegraphics[width=0.24\textwidth]{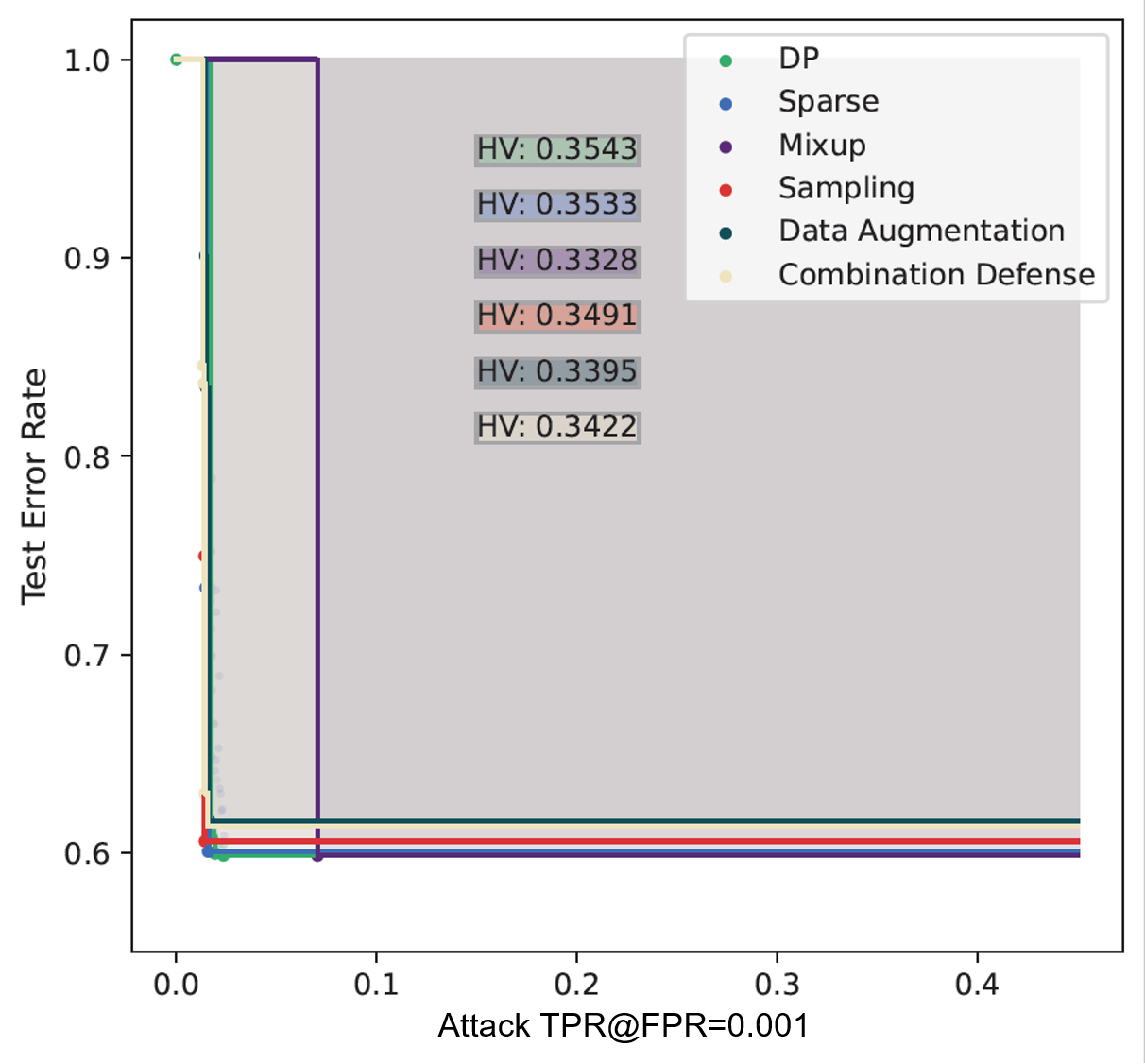} 
    }\hspace{-0.2cm}
    \subfigure[\tiny   AlexNet-CIFAR100 Grad-Cosine]{
        \includegraphics[width=0.24\textwidth]{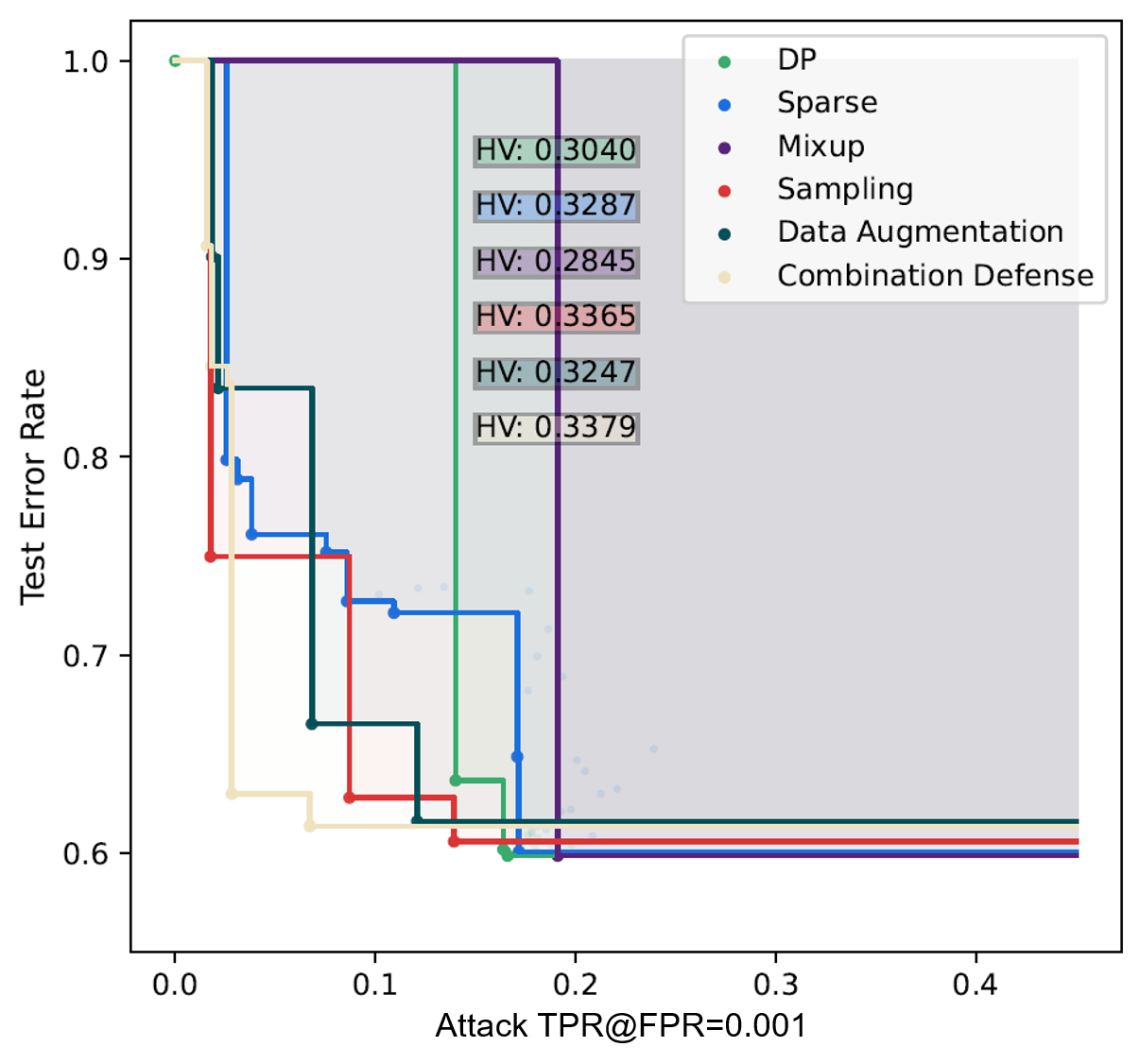} 
    }\hspace{-0.2cm}
    \subfigure[\tiny ResNet18-CIFAR100 Blackbox-Loss]{
        \includegraphics[width=0.24\textwidth]{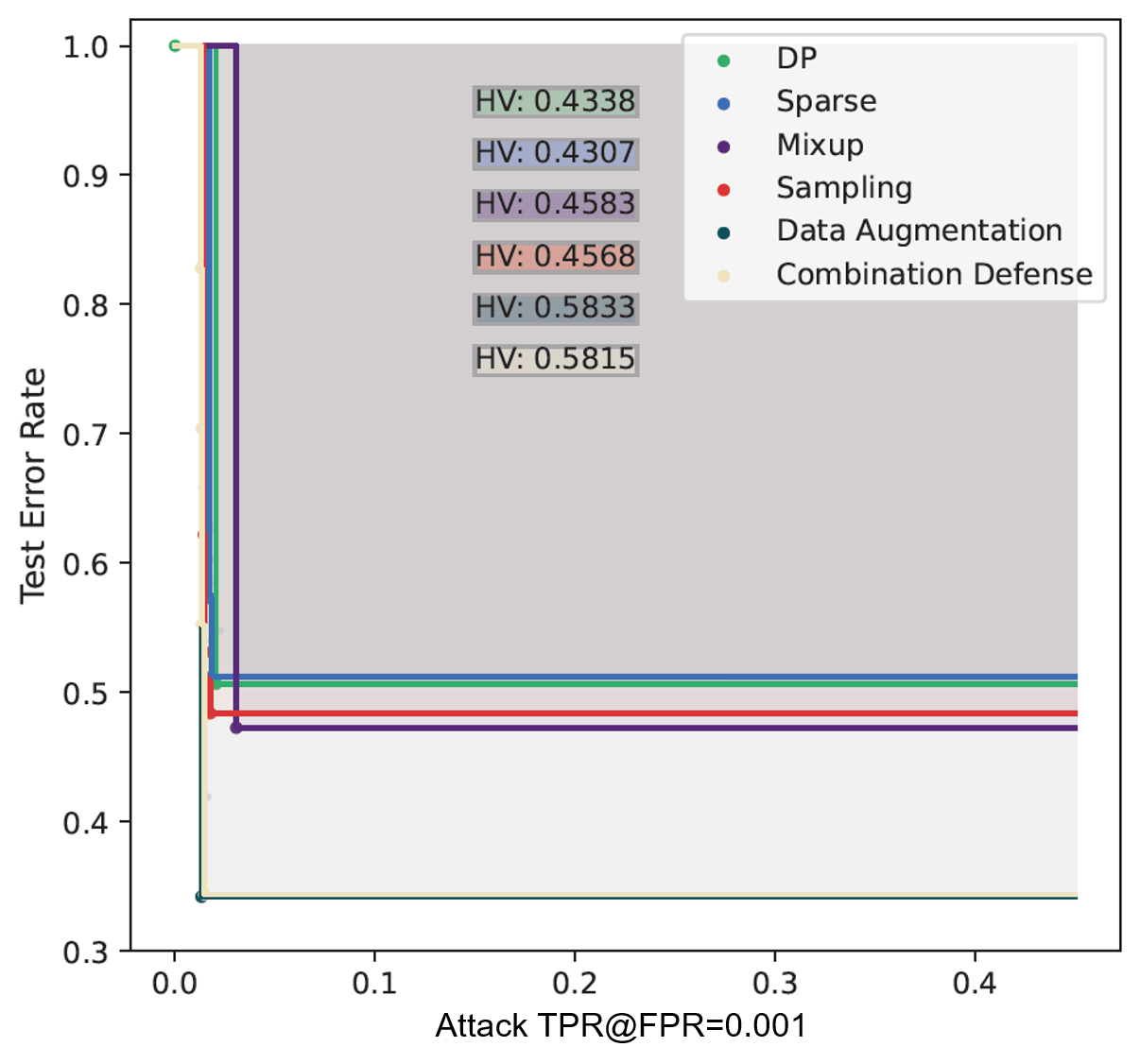} 
    }\hspace{-0.2cm}
    \subfigure[\tiny ResNet18-CIFAR100 Grad-Cosine]{
        \includegraphics[width=0.24\textwidth]{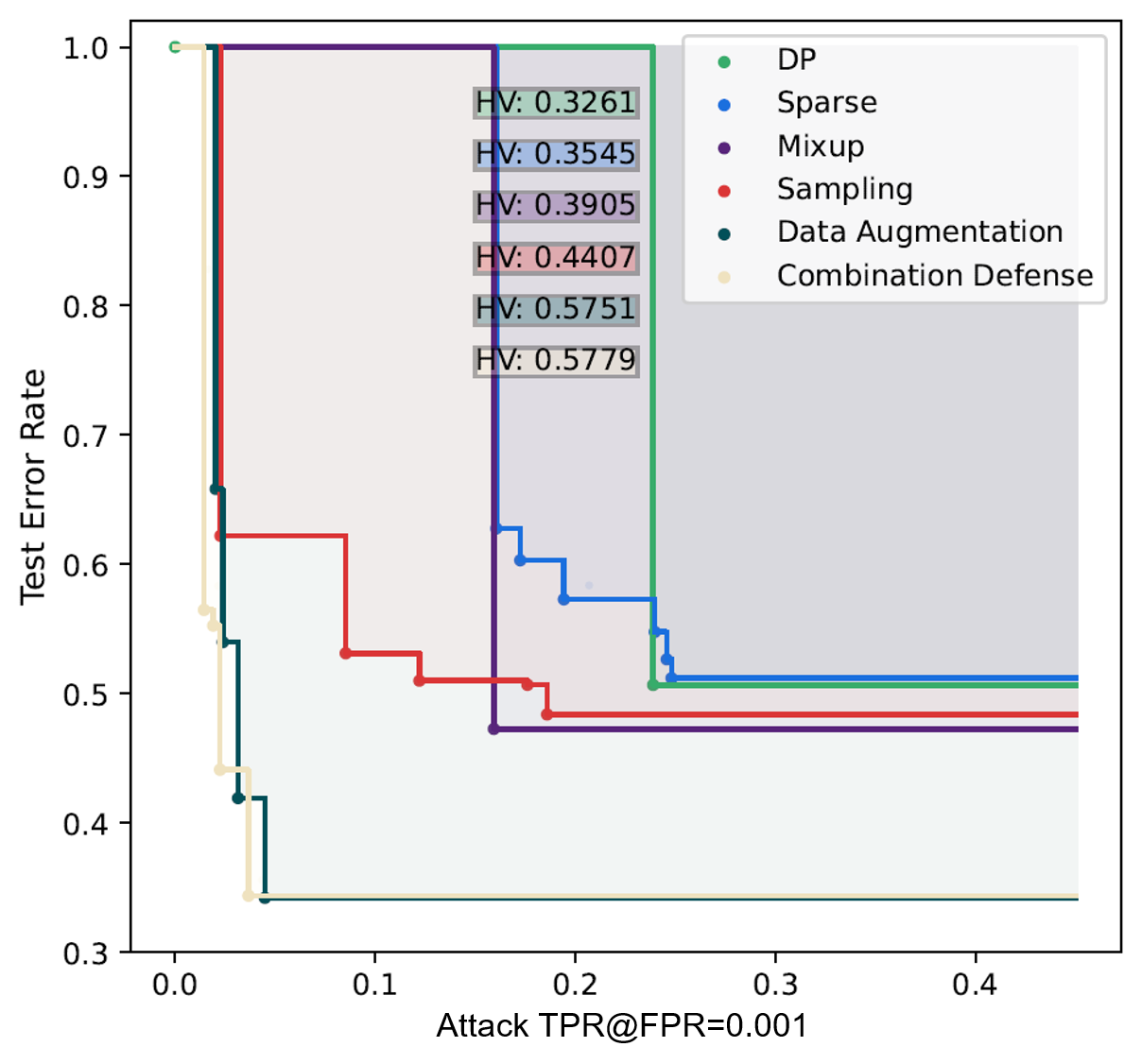} 
    }\hspace{-0.2cm}
        \subfigure[\tiny AlexNet-CIFAR100 Loss-Series]{
        \includegraphics[width=0.24\textwidth]{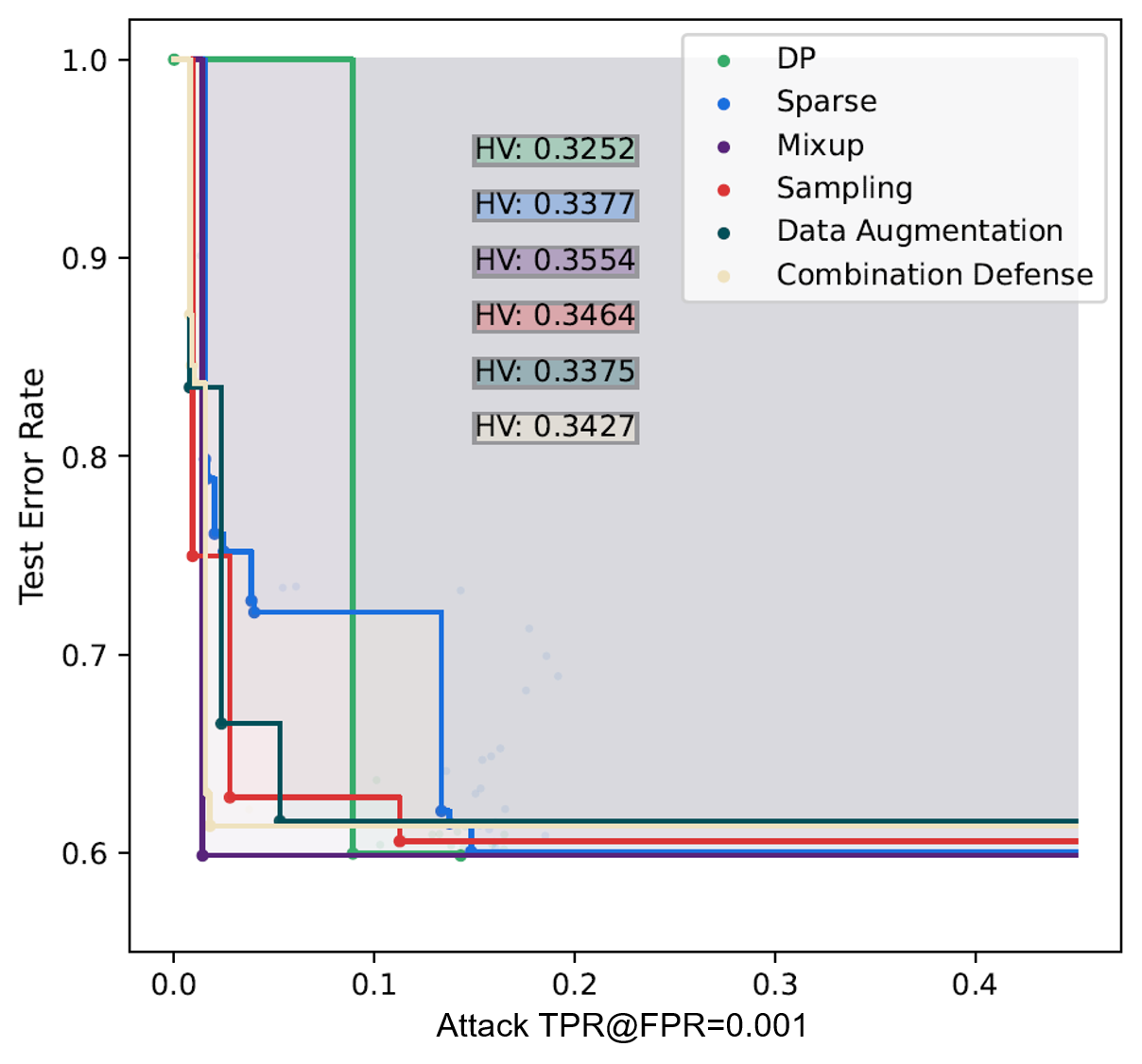} 
    }\hspace{-0.2cm}
    \subfigure[\tiny   AlexNet-CIFAR100 Avg-Cosine]{
        \includegraphics[width=0.24\textwidth]{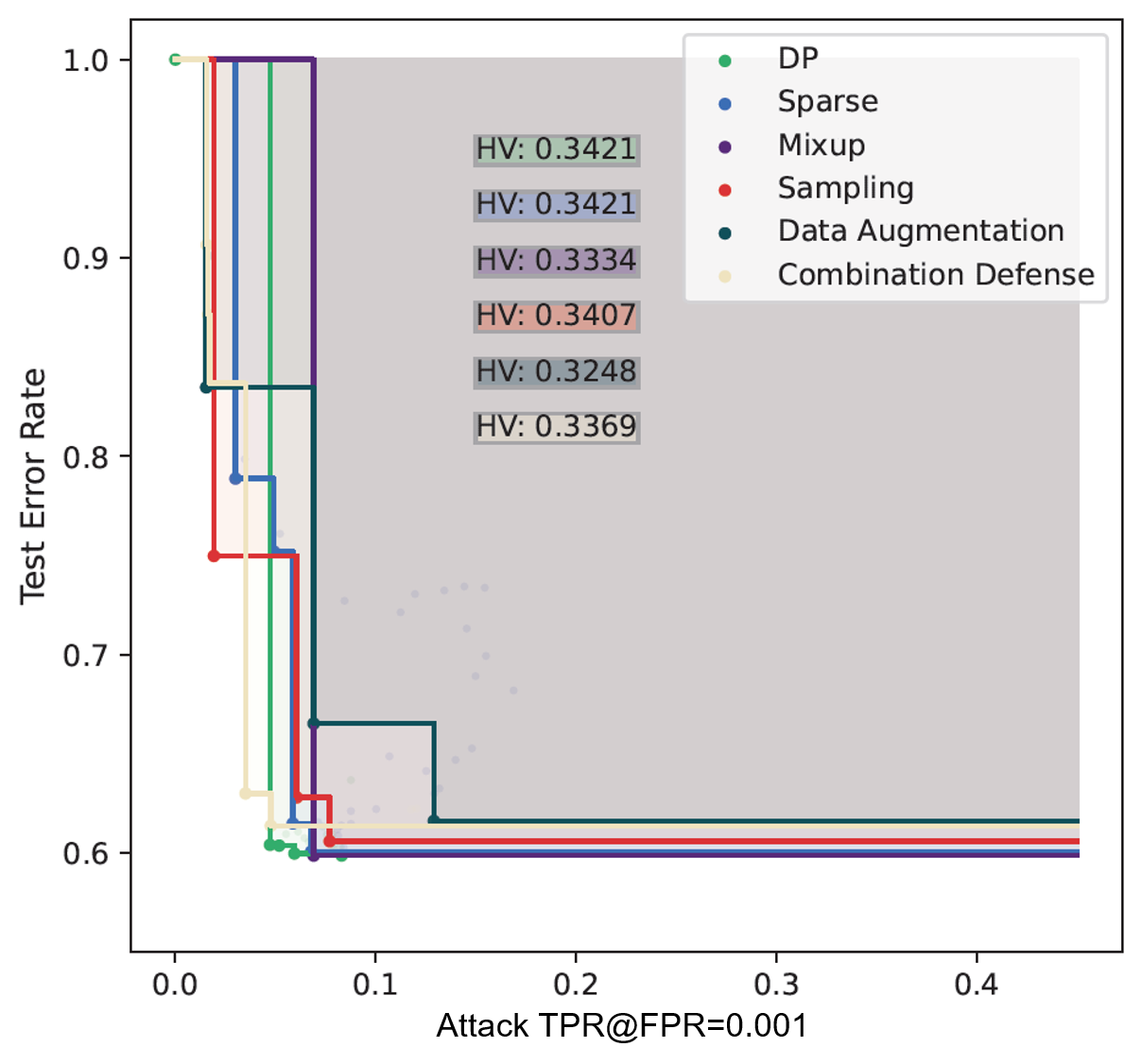} 
    }\hspace{-0.2cm}
    \subfigure[\tiny ResNet18-CIFAR100 Loss-Series]{
        \includegraphics[width=0.24\textwidth]{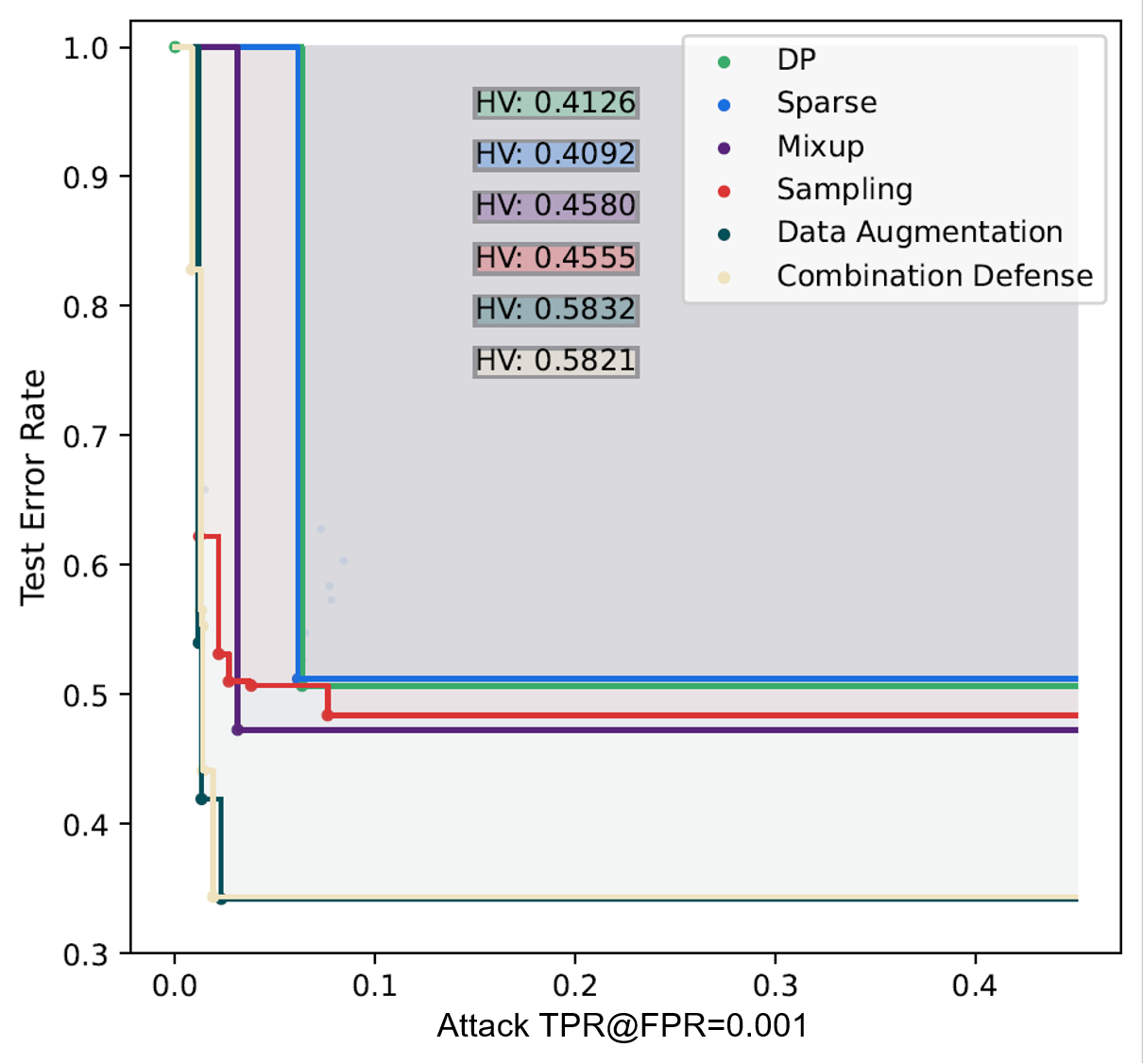} 
    }\hspace{-0.2cm}
    \subfigure[\tiny ResNet18-CIFAR100 Avg-Cosine]{
        \includegraphics[width=0.24\textwidth]{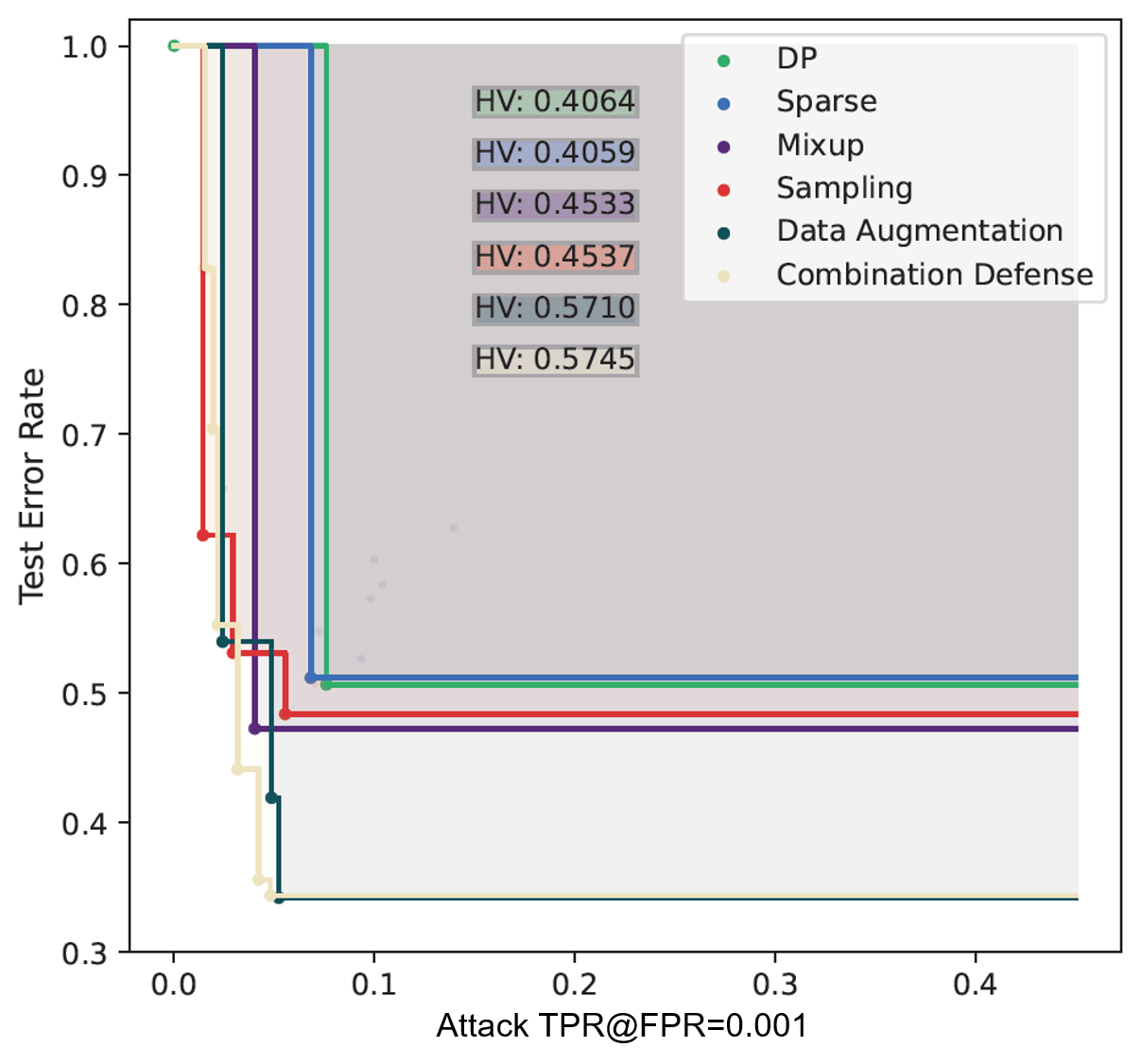} 
    }\hspace{-0.2cm}
        \subfigure[\tiny AlexNet-CIFAR100 FedMIA-I]{
        \includegraphics[width=0.24\textwidth]{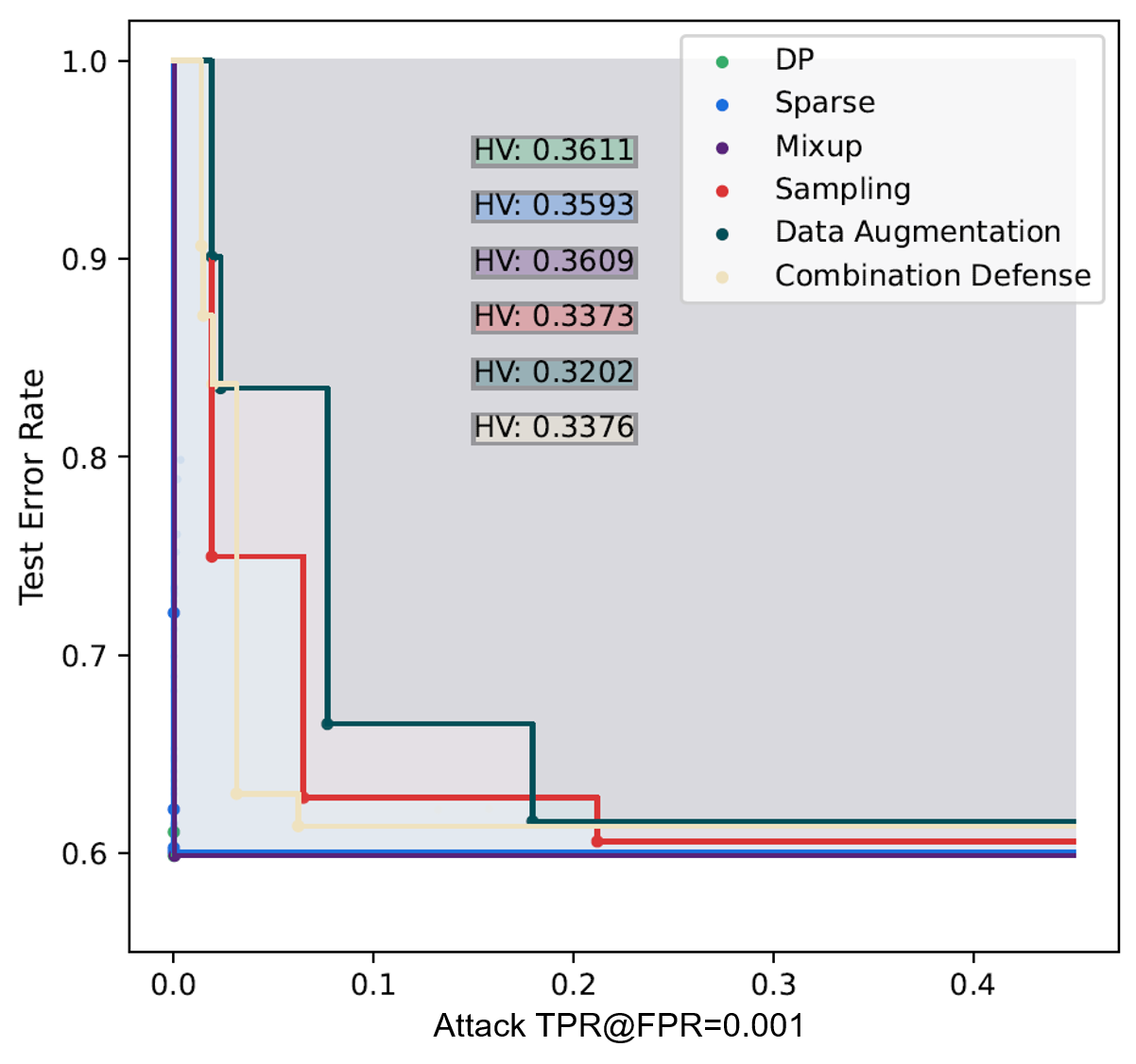} 
    }\hspace{-0.2cm}
    \subfigure[\tiny   AlexNet-CIFAR100 FedMIA-II]{
        \includegraphics[width=0.24\textwidth]{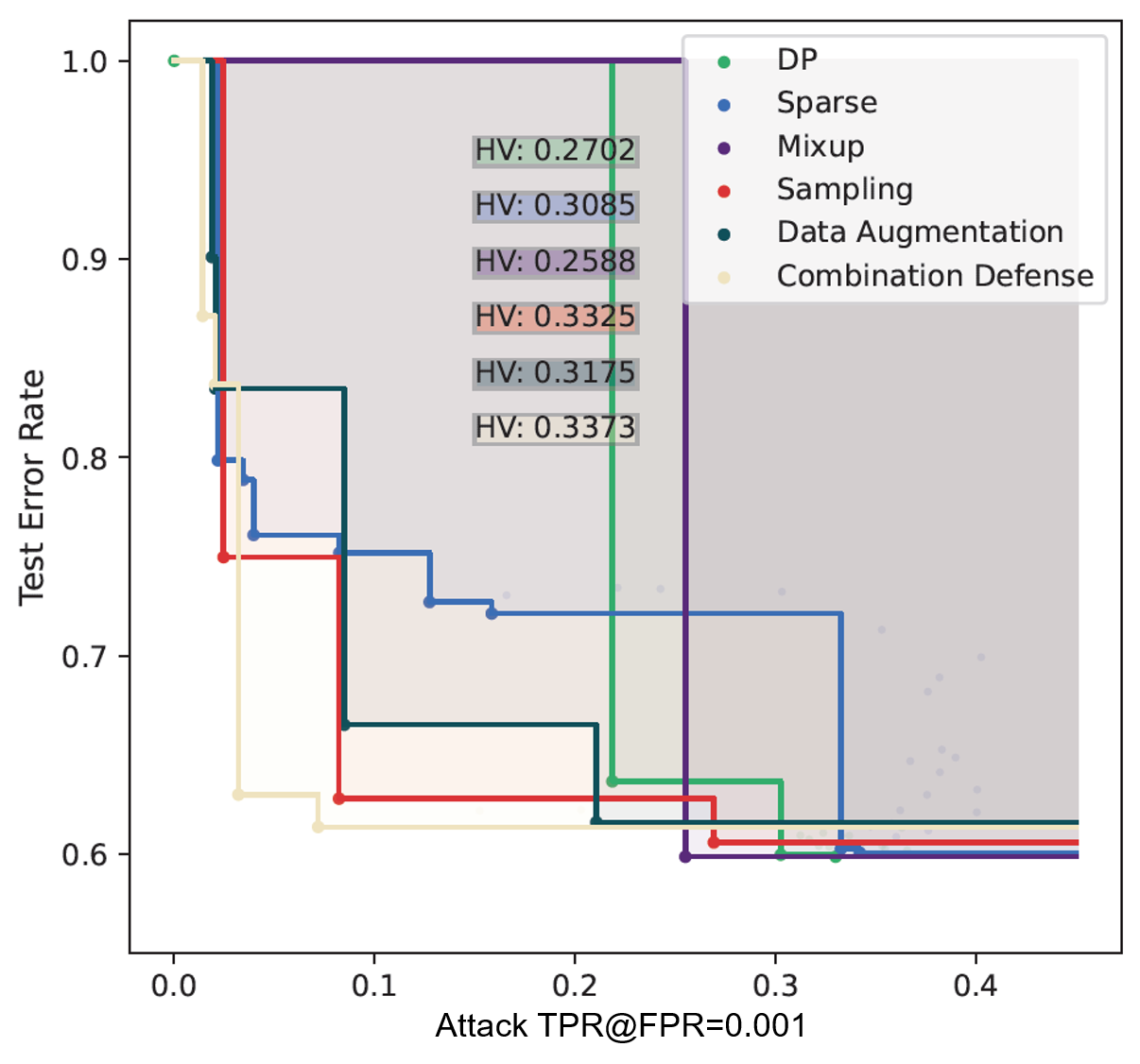} 
    }\hspace{-0.2cm}
    \subfigure[\tiny ResNet18-CIFAR100 FedMIA-I]{
        \includegraphics[width=0.24\textwidth]{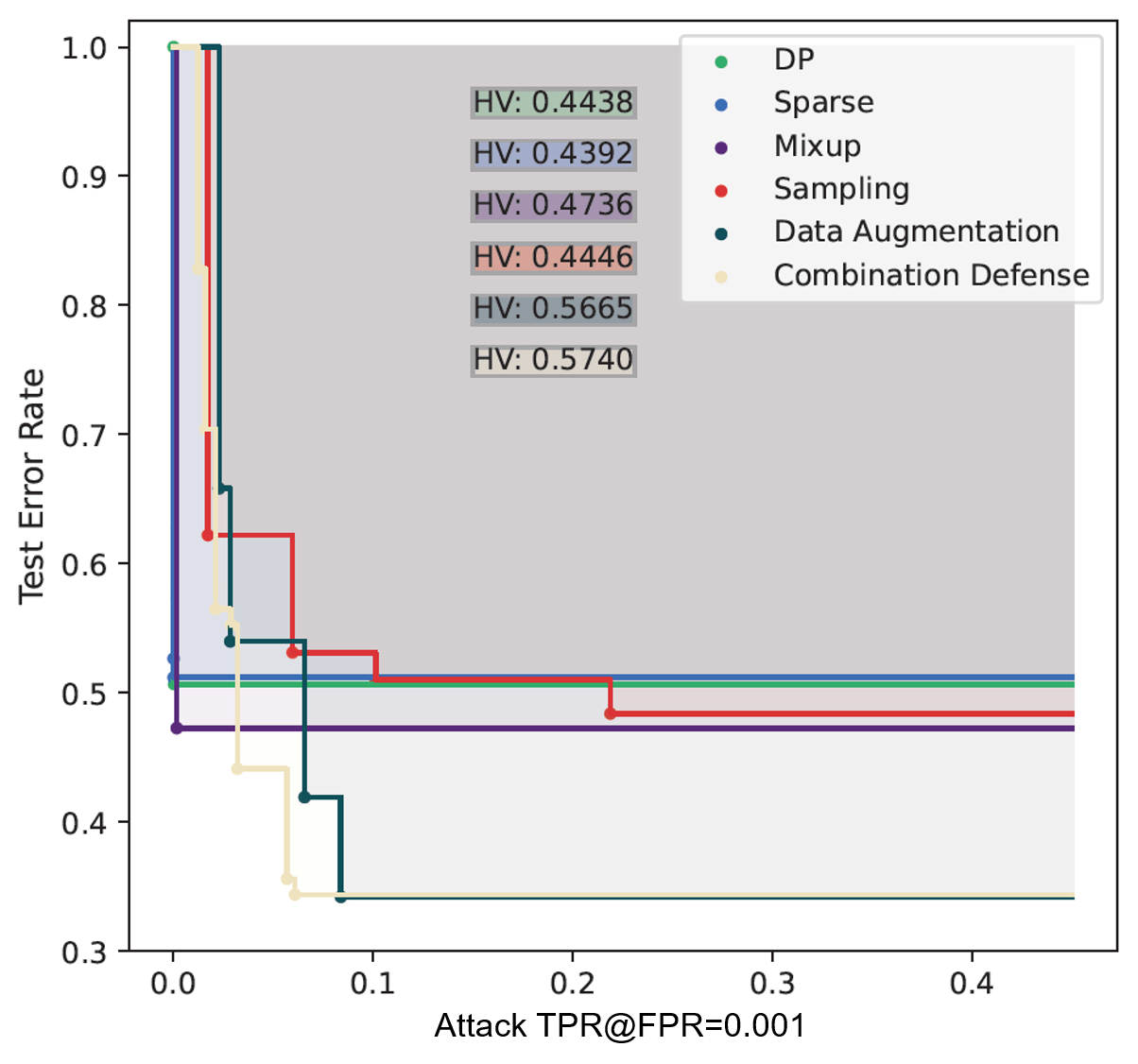} 
    }\hspace{-0.2cm}
    \subfigure[\tiny ResNet18-CIFAR100 FedMIA-II]{
        \includegraphics[width=0.24\textwidth]{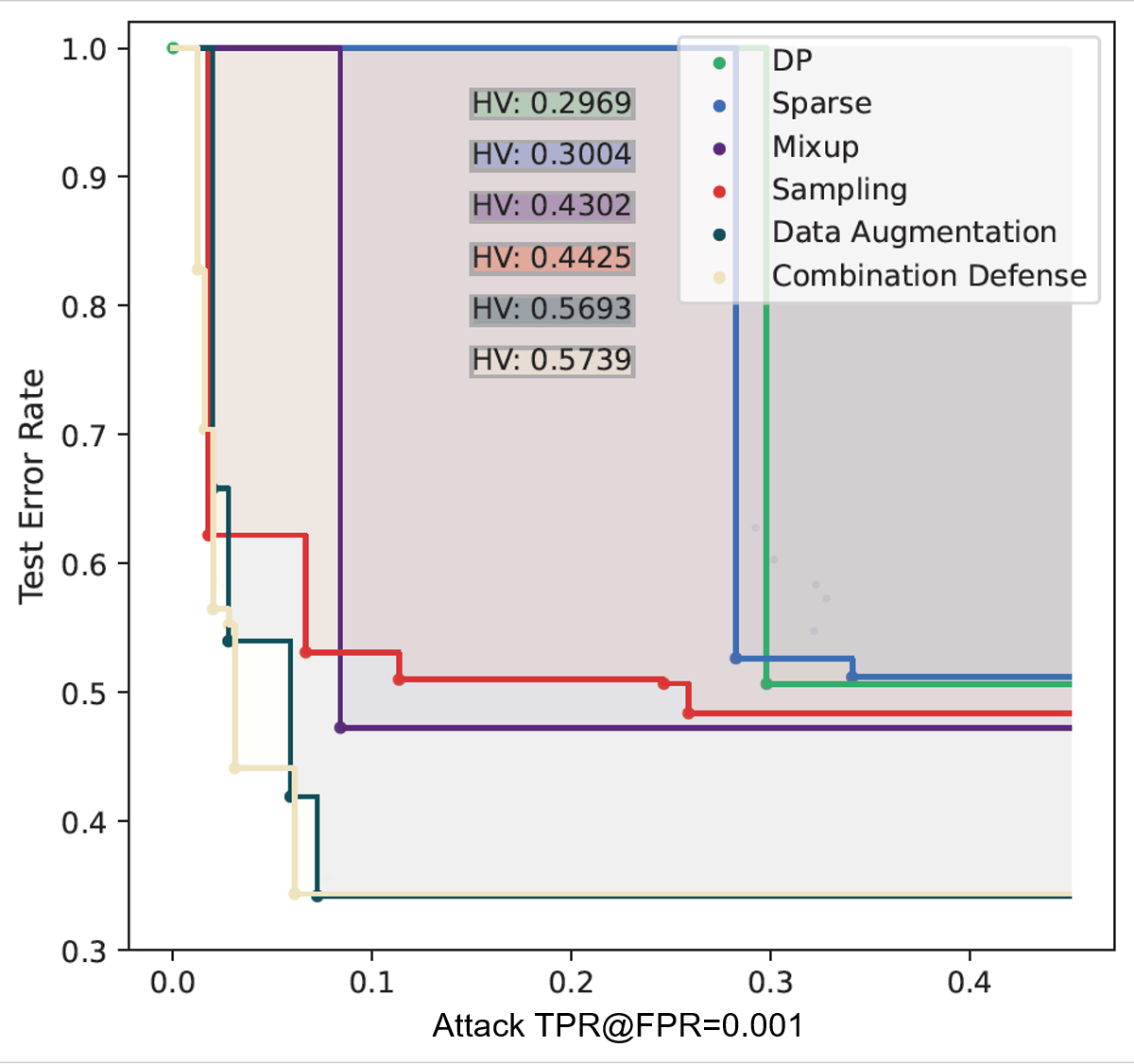} 
    }
    
    \DeclareGraphicsExtensions.
    \caption{Figure (a)-(f) demonstrate the TPR@FPR=0.001 of various defence (including client-level differential privacy (green line) \cite{geyer2017differentially}, sparsification (blue line) \cite{shokri2015privacy}, mixup (purple line) \cite{zhang2017mixup}, data sampling (red line) \cite{li2021sample}, data augmentation (deep blue line) and gradient, combination of data augmentation and sampling (yellow line)) under three attacks (Blackbox-Loss \cite{yeom2018privacy}, Loss-Series \cite{gu2022cs}, FedMIA-I, Grad-Cosine, Avg-Cosine \cite{li2022effective} and FedMIA-II are first, second and third row respectively). A larger hypervolume (HV) \cite{zitzler2004indicator} indicates a better Pareto front of privacy and utility.}
    \label{Fig:def}
    \vspace{-15pt}
\end{figure*}

\subsection{Evaluation Metrics}
\paragraph{\noindent\textbf{Utility loss (Test error rate).}}
In this paper, we quantify the utility loss by using the test error as a metric. The test error measures the accuracy of the model on a separate test dataset, where a lower test error indicates better model utility. The worst possible test error rate is 1, which means that the model makes incorrect predictions for all instances in the test dataset.

\paragraph{\noindent\textbf{Privacy Leakage (AUC and attack TPR).}}
We consider attacks as a binary classification task, and the TPR@FPR of the AUC can be used to measure the accuracy of the classification, which represents the effectiveness of the attack. TPR@low FPR is a metric recently proposed for measuring MIA (Membership Inference Attack). It focuses more on the data that is most susceptible to attacks, and researchers believe that using it as a metric can better characterize privacy protection in worst-case scenarios.

TPR (True Positive Rate) and FPR (False Positive Rate) are two important metrics used to evaluate the performance of binary classification models, such as machine learning algorithms or diagnostic tests. They are calculated as follows:
$$\text{TPR = TP / (TP + FN)}$$
$$\text{FPR = FP / (FP + TN)}$$
where: TP (True Positives) represents the number of positive instances correctly classified as positive.
FN (False Negatives) represents the number of positive instances incorrectly classified as negative.
FP (False Positives) represents the number of negative instances incorrectly classified as positive.
TN (True Negatives) represents the number of negative instances correctly classified as negative.

\paragraph{\noindent\textbf{Hypervolume $HV()$}.} In order to compare Pareto fronts achieved by different defense algorithms, we need to quantify the quality of a Pareto front. To this end, we adopt the hypervolume (HV) indicator \cite{zitzler2004indicator} as the metric to evaluate Pareto fronts. Definition \ref{def:hypervolume} formally defines the hypervolume.

\begin{definition}[Hypervolume Indicator]\label{def:hypervolume}
Let $z = \{z_1,\cdots, z_m\}$ be a reference point that is an upper bound of the objectives $Y = \{y_1,\ldots, y_m\}$, such that $y_{i} \leq z_i$, $\forall i \in [m]$. the hypervolume indicator $\text{HV}_z(Y)$ measures the region between $Y$ and $z$ and is formulated as:
\begin{equation}
    \text{HV}_z(Y) = \Lambda \left( \left \{ q \in \mathbb{R}^m \big| q \in \prod_{i=1}^{m}[y_i, z_i]  \right \}\right) 
   % \text{HV}(\mathcal{X}) = \Lambda \left(  \bigcup_{i \in I} \{ q \big|[f_i, r_i] \}  \right)
\end{equation}
where $\Lambda(\cdot)$ refers to the Lebesgue measure.
\end{definition}

We set the reference point $z$ of privacy leakage and utility loss to be 1 and 100\% respectively.

\section{More experiment}

Figure \ref{Fig:def} demonstrates the privacy-utility tradeoff against different attacks under various defense methods.

\section{Proof of Theorem 1}

\begin{theorem}\label{thm:1-app}
Given the threshold $\delta$, let $\VV_t$ be the member sets estimated by $\hat\Lambda^t$  and $\delta$ in communication round $t$. Let $\Tilde\VV$ be the member sets estimated by $\Tilde\Lambda$ and $\delta$. Then we have
\begin{equation}
    \Tilde\VV  \subset (\VV_1 \cup \cdots \cup \VV_T).
\end{equation}
\end{theorem} 

\begin{proof}
We employ proof by contradiction to establish the theorem. Assume there exists an element \( v \in \mathcal{V} \) such that \( v \notin (\mathcal{V}_1 \cup \cdots \cup \mathcal{V}_T) \).  

By definition, the set \( \mathcal{V}_t \) is defined as  
\[
\mathcal{V}_t = \{v \mid v \in \mathcal{V}, \, \hat{\Lambda}_t(v) \geq \delta \},
\]  
which represents the set of members determined by Eq. (4) in the main text. Additionally, let \( \mathcal{V}_t^n \) denote the non-member set determined by Eq. (4).  

Now, consider an element \( v \notin (\mathcal{V}_1 \cup \cdots \cup \mathcal{V}_T) \). This implies that \( \hat{\Lambda}_t(v) < \delta \) for all \( t \). Consequently, we have:  
\[
\Tilde{\Lambda}(v) = \frac{1}{T} \sum_{t=1}^T \hat{\Lambda}_t(v) < \delta.
\]  
This result indicates that \( v \notin \mathcal{V} \), which contradicts the assumption that \( v \in \mathcal{V} \).  

Thus, the assumption leads to a contradiction, and the proof is complete.

\end{proof}

\end{document}